\documentclass[manuscript, screen, review=False, table,xcdraw]{acmart}
\usepackage{algpseudocode}
\usepackage{algorithm}
\usepackage{xcolor}
\usepackage{multirow}
\usepackage{colortbl} 

\newcommand{\red}[1]{\textcolor{black}{#1}}
\AtBeginDocument{%
  }

\begin{document}

%%
%% The "title" command has an optional parameter,
%% allowing the author to define a "short title" to be used in page headers.
\title{Multi-OCT-SelfNet: Integrating Self-Supervised Learning with Multi-Source Data Fusion for Enhanced Multi-Class Retinal Disease Classification}

%
% The "author" command and its associated commands are used to define
% the authors and their affiliations.
% Of note is the shared affiliation of the first two authors, and the
% "authornote" and "authornotemark" commands
% used to denote shared contribution to the research.
\author{Fatema-E- Jannat}
% \authornote{Both authors contributed equally to this research.}
\email{fjannat@charlotte.edu}
% \orcid{1234-5678-9012}
% \author{G.K.M. Tobin}
% \authornotemark[1]
% \email{webmaster@marysville-ohio.com}
\affiliation{%
  \institution{University of North Carolina at Charlotte}
  \streetaddress{9201 University City Blvd}
  \city{Charlotte}
  \state{North Carolina}
  \country{USA}
  \postcode{28223}
}

\author{Sina Gholami}
\affiliation{%
  \institution{University of North Carolina at Charlotte}
  \streetaddress{9201 University City Blvd}
  \city{Charlotte}
  \state{North Carolina}
  \country{USA}
  \postcode{28223}
  }
\email{sgholami@charlotte.edu}

\author{Jennifer I. Lim, MD}
\affiliation{%
  \institution{University of Illinois at Chicago}
  \city{Chicago}
  \state{IL}
  \country{USA}
  }

\author{Theodore Leng, MD}
\affiliation{%
  \institution{Stanford University School of Medicine}
  \city{Stanford}
  \state{CA}
  \country{USA}
  }

\author{Minhaj Nur Alam}
\email{minhaj.alam@charlotte.edu}
\affiliation{%
  \institution{University of North Carolina at Charlotte}
  \streetaddress{9201 University City Blvd}
  \city{Charlotte}
  \state{North Carolina}
  \country{USA}
  \postcode{28223}
}

\author{Hamed Tabkhi}
\email{htabkhiv@charlotte.edu}
\affiliation{%
 \institution{University of North Carolina at Charlotte}
  \streetaddress{9201 University City Blvd}
  \city{Charlotte}
  \state{North Carolina}
  \country{USA}
  \postcode{28223}
  }

%%
%% By default, the full list of authors will be used in the page
%% headers. Often, this list is too long, and will overlap
%% other information printed in the page headers. This command allows
%% the author to define a more concise list
%% of authors' names for this purpose.
% \renewcommand{\shortauthors}{Trovato et al.}

%%
%% The abstract is a short summary of the work to be presented in the
%% article.
\begin{abstract}
  In the medical domain, acquiring large datasets poses significant challenges due to privacy concerns. Nonetheless, the development of a robust deep-learning model for retinal disease diagnosis necessitates a substantial dataset for training. The capacity to generalize effectively on smaller datasets remains a persistent challenge. The scarcity of data presents a significant barrier to the practical implementation of scalable medical AI solutions. To address this issue, we’ve combined a wide range of data sources to improve performance and generalization to new data by giving it a deeper understanding of the data representation from multi-modal datasets and developed a self-supervised framework based on large language models (LLMs), SwinV2 to gain a deeper understanding of multi-modal dataset representations, enhancing the model's ability to extrapolate to new data for the detection of eye diseases using optical coherence tomography (OCT) images. We adopt a two-phase training methodology, self-supervised pre-training, and fine-tuning on a downstream supervised classifier. An ablation study conducted across three datasets employing various encoder backbones,  without data fusion, with low data availability setting, and without self-supervised pre-training scenarios, highlights the robustness of our method. Our findings demonstrate consistent performance across these diverse conditions, showcasing superior generalization capabilities compared to the baseline model, ResNet-50.

\end{abstract}

%%
%% The code below is generated by the tool at http://dl.acm.org/ccs.cfm.
%% Please copy and paste the code instead of the example below.
%%


% \begin{CCSXML}
% <ccs2012>
%  <concept>
%   <concept_id>00000000.0000000.0000000</concept_id>
%   <concept_desc>Do Not Use This Code, Generate the Correct Terms for Your Paper</concept_desc>
%   <concept_significance>500</concept_significance>
%  </concept>
%  <concept>
%   <concept_id>00000000.00000000.00000000</concept_id>
%   <concept_desc>Do Not Use This Code, Generate the Correct Terms for Your Paper</concept_desc>
%   <concept_significance>300</concept_significance>
%  </concept>
%  <concept>
%   <concept_id>00000000.00000000.00000000</concept_id>
%   <concept_desc>Do Not Use This Code, Generate the Correct Terms for Your Paper</concept_desc>
%   <concept_significance>100</concept_significance>
%  </concept>
%  <concept>
%   <concept_id>00000000.00000000.00000000</concept_id>
%   <concept_desc>Do Not Use This Code, Generate the Correct Terms for Your Paper</concept_desc>
%   <concept_significance>100</concept_significance>
%  </concept>
% </ccs2012>
% \end{CCSXML}

% \ccsdesc[500]{Do Not Use This Code~Generate the Correct Terms for Your Paper}
% \ccsdesc[300]{Do Not Use This Code~Generate the Correct Terms for Your Paper}
% \ccsdesc{Do Not Use This Code~Generate the Correct Terms for Your Paper}
% \ccsdesc[100]{Do Not Use This Code~Generate the Correct Terms for Your Paper}

%%
%% Keywords. The author(s) should pick words that accurately describe
%% the work being presented. Separate the keywords with commas.

% \received{20 February 2007}
% \received[revised]{12 March 2009}
% \received[accepted]{5 June 2009}

%%
%% This command processes the author and affiliation and title
%% information and builds the first part of the formatted document.
\maketitle

\section{Introduction}

In recent days, the artificial intelligence domain has witnessed a revolutionary breakthrough. However, in the medical field, a significant gap persists due to the scarcity of data. Machine learning models require extensive datasets for effective training, yet the medical domain faces constraints in this regard, primarily due to privacy concerns surrounding patient data. This scarcity poses a substantial challenge, hindering the progress and application of scalable medical AI solutions in healthcare.

\begin{figure*}[h!]
\centerline{\includegraphics[width=.80\linewidth]{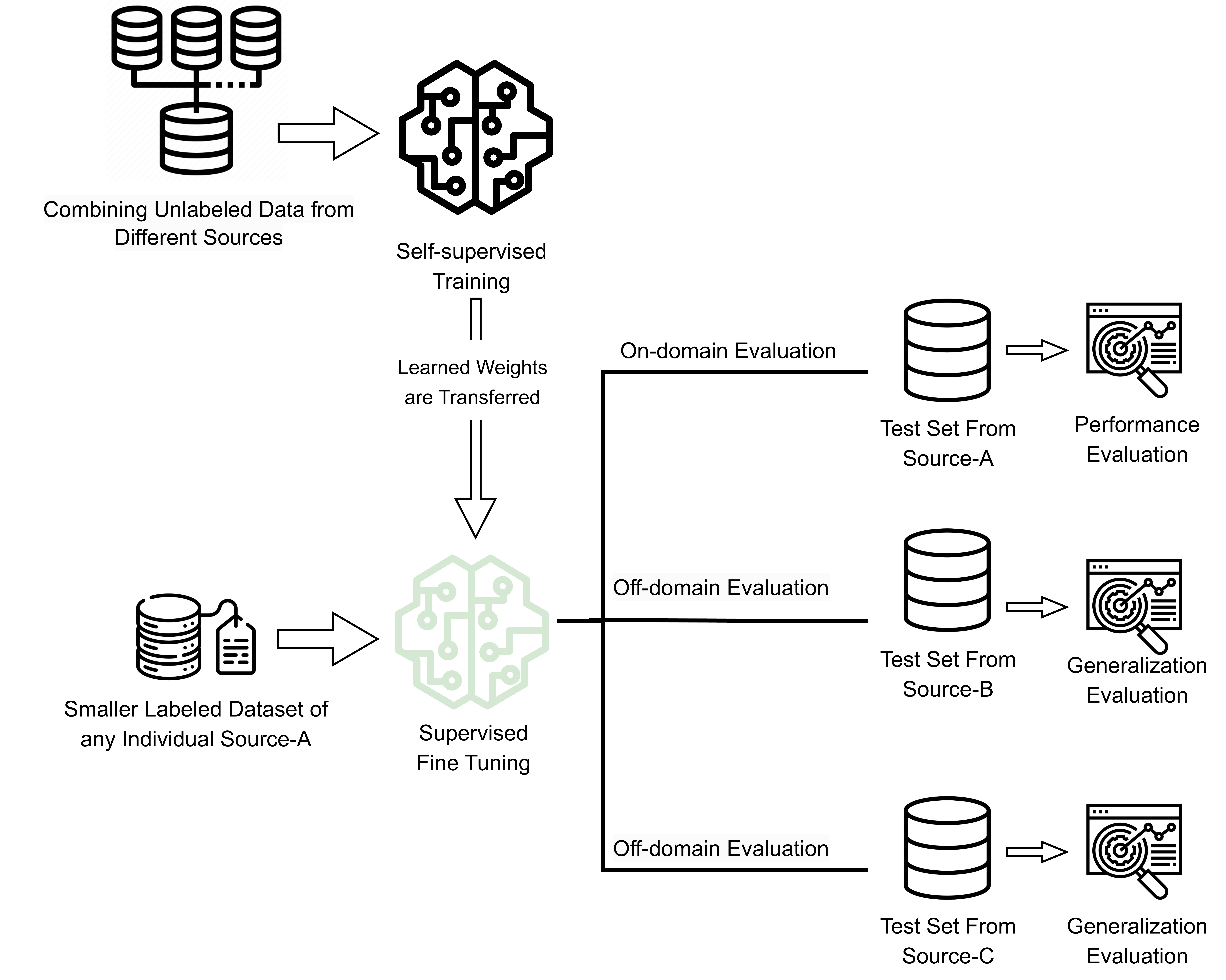}}
\caption{A graphical depiction of our methodology, consolidating fundamental concepts and procedural steps.}
\label{fig-intro}
\end{figure*}

To bridge this gap, our research addresses two key challenges. First, we focus on developing a robust machine learning classifier based on Large Language Models (LLM) to detect eye diseases from optical coherence tomography (OCT) images for AI-based eye care management. Second, we address the challenge of creating a machine learning model capable of learning from varied unlabeled data, making it useful in real-world situations with new and unseen data.

Age-related macular degeneration (AMD), along with other sight-threatening conditions such as diabetic macular edema (DME), choroidal neovascularization (CNV), and diabetic retinopathy (DR), ranks among the leading causes of irreversible blindness and vision impairment (VI) globally. VI affects nearly 2.2 billion people worldwide, with almost 1 billion cases potentially preventable through early diagnosis and intervention \cite{who_blindness_visual_impairment}. Therefore, it is critical to identify those who are at risk of developing the disease or seeing it progress, especially from the early stages to the more advanced stages, as prompt intervention can stop the disease's progression or slow it down, ultimately preventing irreversible VI. Individuals at high risk for VI would greatly benefit from more frequent ophthalmic examinations, continuous monitoring, and prompt treatment \cite{scott2013long}.  Leveraging AI-based tools for early detection and continuous monitoring could significantly enhance our ability to identify at-risk individuals and intervene promptly, potentially saving countless individuals from needlessly suffering vision loss and impairment \cite{yi2009spectral}.

Automating diagnosis in ophthalmology has shown great promise with machine learning (ML) and deep learning (DL) techniques \cite{friberg2011analysis, alam2020quantitative, schmidt2018prediction, wang2016genetic}. However, the limited diversity of training datasets frequently impedes their effectiveness in the application of real-world clinical settings. When implementing these models in clinical settings, it is important to make a variety of datasets sourced from multiple institutions accessible to optimize their usefulness in clinical workflows. These datasets use different OCT image-capturing devices, cover a variety of demographics, and follow different protocols. By exposing our models to a range of datasets, we can increase their scalability, versatility, and adaptability, which will ultimately improve their performance and usefulness in real-world clinical scenarios.

This work improves retinal imaging detection and diagnosis, especially in automated ophthalmic diagnosis, by utilizing recent advances in large pre-trained transformer networks. The evolution of transformer models from Natural Language Processing (NLP) to the computer vision domain underscores their capability and scalability. The transformer model, first introduced in 2017 by Vaswani et al \cite{vaswani2017attention} for the NLP task leverages the self-attention mechanism to learn the contextual relationship among words within a sentence. This innovative approach revolutionized the NLP by enabling advancements in critical tasks like text summarization, translation, and sentence completion. Transformer model BERT \cite{devlin2018bert}, introduced in 2018, and trained on an extensive corpus of text data, stands as one of the most popular models capable of performing diverse tasks in NLP.
This breakthrough performance of transformer models in NLP has raised great interest in the computer vision community. Since this architecture can learn the long-range dependencies within the data, it allows us to grasp the spatial and temporal relationships within images. By processing images as a sequence of patches, it can also capture the global context which is very important in tasks such as image classification, object detection, and pose estimation. With the introduction of the first image-based transformer architecture, Vision Transformer (ViT)\cite{trans001}, in 2020 by Alexey Dosovitskiy et al., it achieved state-of-the-art performance on image classification tasks, showcasing the potential applications of this transformer architecture in the field of computer vision such as object detection, image segmentation, object tracking, etc. Later several transformer models were developed such as DeiT \cite{touvron2021training}, DETR \cite{carion2020endtoend},  Swin Transformer \cite{liu2021swin}, SwinV2 \cite{liu2022swin}, VisionLLM \cite{wang2023visionllm} for the computer vision domain.
However, employing a transformer architecture requires a significant amount of computational resources, posing a challenge for communities with limited computational infrastructure. Moreover, a large dataset is crucial for effective training, creating problems in cases where such extensive datasets are unavailable. Despite these challenges, researchers are actively working on solutions to address these issues through continuous advancements in hardware, the development of efficient algorithms, the implementation of data augmentation techniques, and the integration of synthetic data.

Inspired by masked autoencoders \cite{he2022masked}, we have developed a large-scale self-supervised model with random masking, utilizing a variation of transformer models the SwinV2 \cite{liu2022swin} backbone which is specifically for the classification of retinal diseases from optical coherence tomography (OCT) images. Our SwinV2-based classifier leverages the transformer architecture, a foundational component also employed in Large Language Models. Its attention mechanism enables the model to dynamically determine the relative importance of various input data regions, allowing it to capture complex patterns and relationships in the data. Similar to how LLMs process and comprehend textual data by concentrating on context and semantic relationships, SwinV2 performs well in visual tasks by paying attention to relevant image regions, which improves the accuracy of feature extraction and classification.  The self-attention mechanism is applied to image patches by the SwinV2 model. SwinV2 is able to model global and local features more effectively by treating images as sequences of patches, in a manner similar to how text is processed as sequences of words or tokens. This leads to enhanced performance in image classification tasks. The integration of SwinV2 in our OCT image classification task demonstrates the versatility of transformer architectures beyond natural language processing. Notably, our focus extends to the multi-class classification task, encompassing the discrimination of normal cases from those presenting with AMD, choroidal neovascularization (CNV), diabetic macular edema (DME), and diabetic retinopathy (DR). Expanding on our earlier study,  "OCT-SelfNet: A Self-Supervised Framework with Multi-Modal Datasets for Generalized and Robust Retinal Disease Detection" \cite{jannat2024octselfnet}  which concentrated solely on binary classification, this work extends the scope to a multi-class problem. 
% To enhance model performance and capability to capture complexities of multi-class datasets, we have incorporated additional layers and heads into the transformer network. 
Through the utilization of self-supervised learning (SSL), our model aims to alleviate the necessity for extensive manual annotations by experts, thereby reducing workload and facilitating broader clinical AI applications within retinal imaging data. Importantly, our model exhibits the capability to learn versatile and generalizable features from unlabeled retinal OCT datasets, a critical aspect for developing AI systems requiring fewer labeled examples to adapt to diverse diagnostic tasks.

Our study stands in contrast to previous research \cite{leandro2023oct, tsuji2020classification, LEE2017322, 8120661, lu2018deep} which have primarily focused on the analysis of singular datasets in isolation. In this methodology the training centers on individual datasets, where models are trained on specific segments and then evaluated on the remainder. However, this type of framework presents challenges, particularly in practical clinical settings with limited dataset sizes. Deep learning models require larger datasets for effective training, and smaller datasets often result in poorer accuracy. Moreover, deploying a model trained on one clinical dataset to another setting is problematic due to variations in device settings and environmental factors. To handle these challenges, we investigate a more intricate method by examining complexities posed by domain adaptation across multiple datasets. We integrate several OCT datasets—DS1, DS2, and DS3—sourced from three distinct studies conducted by Kermany et al. \cite{kermany}, Srinivasan et al. \cite{srinivasan}, and Li et al.\cite{li2020octa}, respectively. 

\red{This approach was motivated by several key considerations. First, integrating multiple datasets will allow the pre-trained model to learn from a more diverse dataset with different clinical settings, capturing a broader spectrum of imaging conditions. This diversity will help to develop a robust model that will generalize well across unseen patient populations and clinical settings. Second, the combined dataset will increase the amount of available training data. A larger dataset enables more effective training, as the model can be exposed to a wider variety of examples, reducing the risk of overfitting. Third, pre-training the model on this comprehensive dataset will allow it to leverage transfer learning effectively. The weights learned during the pre-training stage will serve as a strong foundation for the classifier network during fine-tuning. By starting with a model already familiar with a wide range of visual features, we can achieve better performance with less training data specific to our final task, improving both training efficiency and final accuracy.}

\red{This comprehensive fusion of datasets allows us to construct a unified dataset, leveraging insights and data from various modalities to develop a more holistic understanding of domain adaptation complexities.}

\red{As a result of our approach, our model’s ability to generalize to novel, unseen data is enhanced. This proves especially beneficial in situations where access to extensive datasets is limited. Within the self-supervised pre-training phase, the model undergoes the training and validation processes on a combined large dataset. During this stage, we leverage the combined training and validation datasets from DS1, DS2, and DS3 to ensure comprehensive learning. Following pre-training, the model proceeds to the fine-tuning phase, where it is individually fine-tuned on each dataset. This fine-tuning process allows the model to adapt its learned representations to the specific characteristics of each dataset. Subsequently, the model is cross-evaluated across all test sets to assess its robustness and effectiveness across different datasets.}

Our proposed method follows a comprehensive process that encompasses several key stages. Initially, the data fusion process occurs during the self-supervised pre-training phase, where information from multiple datasets is integrated to enhance model understanding and performance. Subsequently, fine-tuning is conducted on individual datasets to tailor the model to specific domain characteristics and optimize its performance further. Following fine-tuning, evaluation takes place on respective test sets, providing on-domain assessment measurements. Additionally, the model's generalization capability is evaluated on other test sets for off-domain evaluation to measure its performance across diverse datasets and scenarios, thereby assessing its robustness and generalization capability. This process as depicted in Figure \ref{fig-intro}, illustrates the systematic approach employed in our methodology.

Comparative analysis against the baseline model, ResNet-50, offers valuable insights into the performance of our proposed framework, Multi-OCT-SelfNet. Utilizing ResNet-50 as a benchmark, we trained it on individual datasets and conducted cross-evaluation across all test sets to establish a reference point for comparison. Through an ablation study, we consistently observed the superiority of our Multi-OCT-SelfNet framework. AUC-ROC (Area Under the Receiver Operating Characteristic curve) and AUC-PR (Area Under the Precision-Recall curve) values are specifically used to measure performance. When comparing our suggested framework to the baseline ResNet-50 model, these metrics provide strong indications of its efficacy and dependability.

The primary contributions of this paper are outlined as follows:

\begin{itemize}
    \item This work presents an approach of combining multiple datasets and utilizing different modalities and showcases its efficacy in significantly improving classification performance on unseen datasets, while also demonstrating robust domain generalization capabilities.
    
    \item \red{This paper introduces a two-phase methodology: firstly, employing a SwinV2-based masked autoencoder during pre-training, followed by a fine-tuning stage classifier for the classification of retinal diseases, specifically designed for Optical Coherence Tomography (OCT) use cases and multi-class classification tasks.}

    \item Extensive evaluation and ablation studies conducted in this paper illustrate the robustness and generalization capabilities of the proposed approach. Remarkably, even across different test sets this method exhibits improved performance without additional fine-tuning. Such findings have promising implications for the integration into real-world clinical settings.
    
    \end{itemize} 

\section{Related Works}

In recent years, computer vision has emerged as an important tool in medical imaging, facilitating advanced diagnostics, treatment planning, and disease monitoring. The integration of deep learning (DL) techniques has particularly revolutionized this field, by automating tasks such as image classification, segmentation, and disease diagnosis, thereby revolutionizing medical image analysis. Concurrently, transformer networks, originally designed for natural language processing, have shown promising potential, broadening the scope of applications within medical imaging.  This section explores the latest developments and research efforts in leveraging DL and transformer networks for medical image analysis, with an emphasis on the contributions and advances made by them.

%DL in Medical Field

The field of computer vision has undergone a significant transformation due to the evolution of deep learning (DL), which began with the introduction of AlexNet \cite{NIPS2012_c399862d} in 2012. Serving as one of the first deep convolutional neural network (CNN) models, AlexNet's success in the ImageNet was a major turning point in computer vision methodologies, transitioning from traditional methods to automated DL techniques. Subsequently, numerous backbone models like VGG \cite{vgg}, ResNet \cite{resnet}, and Inception \cite{inception} have developed, further advancing image analysis capabilities. These advancements extended beyond traditional computer vision, impacting medical domains by demonstrating effectiveness in tasks such as medical image classification \cite{shazia2021comparative, krishnapriya2023pre, srinivas2022deep, bressem2020comparing, yang2021detection, choi2021deep, 10.1167/tvst.9.2.35, Alam:20}. Notably, DL techniques have found significant application in the classification of optical coherence tomography (OCT) images, particularly in diagnosing conditions like age-related macular degeneration (AMD), choroidal neovascularization (CNV), and diabetic macular edema (DME). Studies such as those referenced by \cite{tsuji2020classification} and \cite{leedeep} have underscored the remarkable accuracy and effectiveness of DL in distinguishing abnormal OCT images from normal ones, indicating the potential for automated screening and the development of computer-aided diagnostic tools. However, DL algorithms, particularly CNNs, need substantial amounts of training data, which can be challenging to obtain in medical imaging where data scarcity is common. Consequently, techniques such as transfer learning and domain adaptation have become essential for leveraging knowledge from source tasks to enhance performance in target tasks, addressing the data scarcity issue.

%Transformers in Medical Field

In recent advancements within medical image analysis, transformer models have emerged as a promising avenue for enhancing diagnostic accuracy and efficiency. Originally developed for natural language processing tasks, transformers have been adapted to handle the complex spatial relationships present in medical images. 
In 2017 \cite{vaswani2017attention}, the transformer model was first introduced for the NLP task which leverages the self-attention mechanism to learn the contextual relationship among words within a sentence. Drawing inspiration from this concept, the vision transformer (ViT) \cite{trans001} utilizes multi-head attention mechanisms to understand the contextual relationships among image pixels. This design excels in learning long-range dependencies in data, enhancing its ability to interpret spatial and temporal aspects of images. By treating images as sequences of patches, ViT effectively comprehends the overall context, a crucial factor in image classification, object detection, and pose estimation. The introduction of ViT marks a significant milestone, setting new benchmarks in image classification and showcasing its immense potential in various computer vision applications including medical image analysis. The application of transformer models in medical imaging, particularly in ophthalmology, has been extensively studied, as evidenced by works such as Ayana et al. \cite{ayana2023vision}, Okolo et al. \cite{okolo2022ievit}, Alshammari et al. \cite{alshammari2022olive}, Wang et al. \cite{wang2022semi}, and Kihara et al. \cite{kihara2022detection}. Notably, Wu et al. \cite{wu2021vision} proposed a transformer-based approach tailored for fundus image analysis, wherein images are segmented into patches for sequential classification. This method has demonstrated remarkable performance in contrast to traditional convolutional neural networks (CNNs) across various metrics such as accuracy, specificity, precision, sensitivity, and quadratic weighted kappa score. The success of this method underscores the effectiveness of the applicability of attention mechanisms in diagnosing diabetic retinopathy. However, the adoption of a Vision Transformer (ViT) architecture poses challenges due to its heavy computational requirements, as highlighted by Islam et al. \cite{islam2022recent}.

%Related works on SSL
The development of a large language model, named BERT (Bidirectional Encoder Representations from Transformers) \cite{devlin2018bert}, has revolutionized natural language processing tasks, demonstrating the power of self-supervised learning techniques. Inspired by their success in language understanding, researchers have begun exploring the application of these models in the computer vision domain. By leveraging the pre-trained representations learned from vast amounts of data,  these models provide a novel way to address problems in medical image analysis, like limited labeled data scenarios. The subsequent introduction of Masked Image Modeling (MIM) marked a significant advancement in the field of self-supervised learning (SSL). MIM techniques, such as the masked autoencoder (MAE) introduced by He et al. \cite{he2022masked}, focus on reconstructing masked portions of input data, allowing models to learn robust feature representations by capturing the underlying structure of visual data. Building upon this foundation, SimMIM \cite{simmim} introduced a straightforward but successful method that directly predicts the pixel values of masked patches in images. These advancements highlight the potential of MIM-based approaches to address challenges in medical imaging, including data scarcity and the need for robust feature extraction, ultimately advancing diagnostic capabilities.

Several studies, including those by Fang et al. \cite{fang2022self}, Qiu et al. \cite{qiu2019self}, and Jing et al. \cite{jing2020self}, have underscored the growing importance of self-supervised learning (SSL) in the domain of ophthalmology-focused deep learning research. Tang et al. \cite{tang2022selfsupervised} demonstrated the effectiveness of SSL in conjunction with the Swin UNETR architecture for analyzing 3D medical images, achieving state-of-the-art performance. Their findings highlight the potential of SSL to address challenges such as the scarcity of labeled data and the necessity for patient-specific diagnostic tools. 

These works advance our understanding of how transformer and SSL can effectively extract meaningful information from large quantities of unlabeled data, ultimately advancing the capabilities of applicability of scalable medical AI solutions.

\section{Methodology}

In this work, we have taken a holistic approach by combining datasets sourced from three distinct origins. This process adds a wider range of information to the overall representation learning because each dataset contributes distinct modalities. Combining a wide range of data sources improves performance and the model's capacity to generalize to new data by giving it a deeper understanding of the data representation. We trained an SSL MAE network with SwinV2 as its backbone architecture by utilizing this combined dataset. This pre-trained weight makes a robust foundation for classifying retinal diseases effectively in the downstream tasks.

\begin{figure*}[h!t]
\centerline{\includegraphics[width=.95\linewidth]{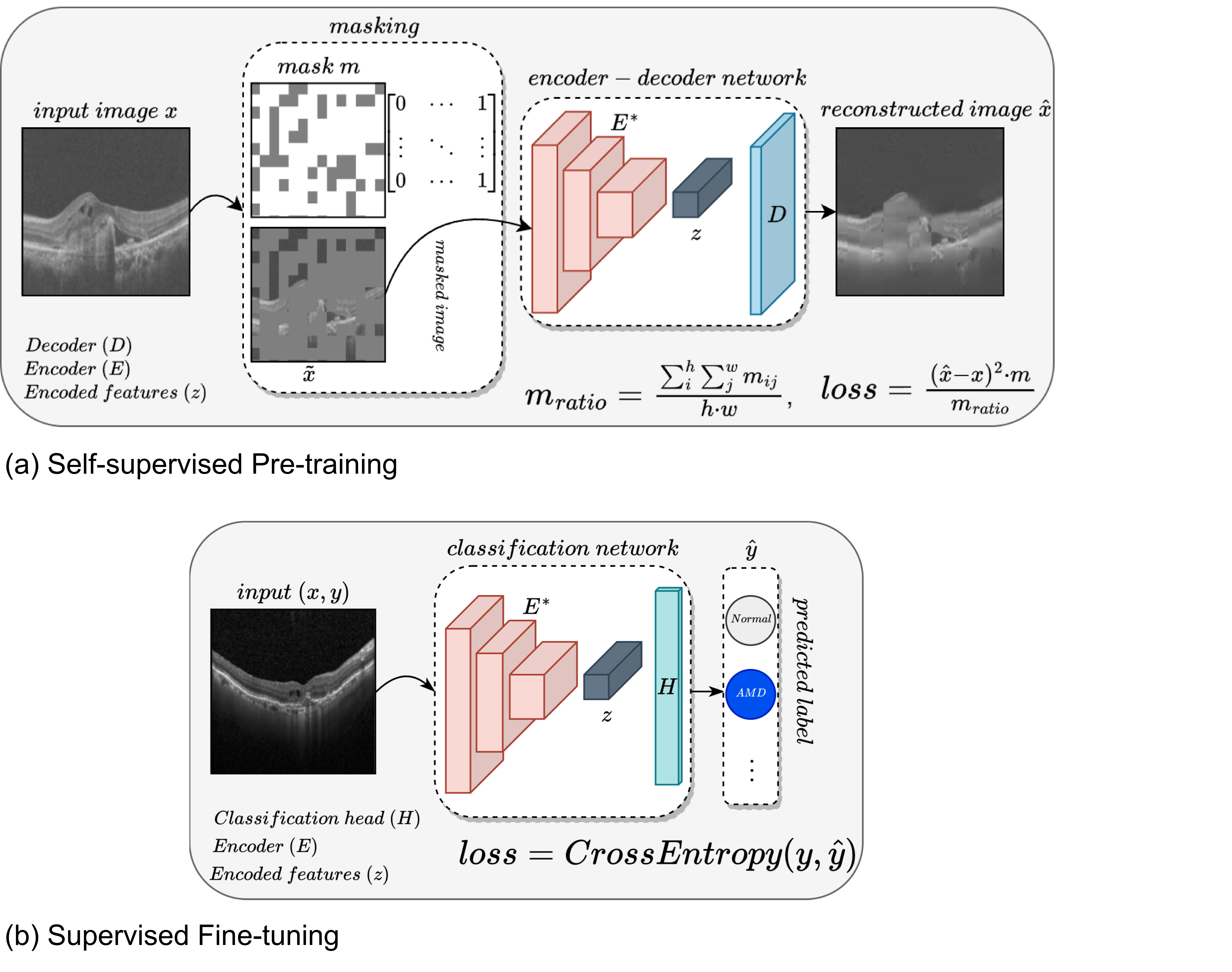}}
\caption{Overview of the Framework: (a) In the initial pre-training phase, the framework utilizes masked image autoencoder as a self-supervised task to learn representations from unlabeled images. In this process, a random subset of image patches is masked and fed into the auto-encoder to reconstruct it. (b) In this phase, the pre-trained encoder from the first phase is employed along with a linear classifier for the classification task. The learned weights from the pre-training phase are transferred to the fine-tuning phase.}
\label{fig-framework}
\end{figure*}

There are four essential stages in our proposed framework. 
\begin{itemize}
\item [](1) Data Fusion: Our study used three OCT image datasets, combining their training and validation sets for Self-Supervised Pre-training, followed by individual fine-tuning on each dataset to evaluate classification performance and generalization.

\item[](2) Self-Supervised Pre-training: In this initial stage of training, a self-supervised pre-training is conducted on a combined collection of unlabeled Optical Coherence Tomography (OCT) images, employing a transformer-based Masked Autoencoder(MAE) approach to extract detailed visual representations. Through this self-supervised learning process, the model gains an understanding of the structure and features within the multimodal OCT images. Subsequently, this learned weight is transferred to a supervised classifier model, leveraging the learned representations to improve the classification task.

\item [](3) Supervised Fine-tuning: Following the self-supervised pre-training phase, a supervised fine-tuning is conducted. This fine-tuning process aims to refine the model's classification capabilities by leveraging the weights transferred from the pre-trained model. By exposing the model to labeled data and adjusting its parameters based on the specific task requirements, the fine-tuning stage further optimizes the model's performance, enhancing its ability to accurately classify retinal diseases from OCT images.

\item [](4) Baseline Training: In our evaluation study, ResNet50 was used as the baseline model against which we compared the performance of our proposed model. By employing ResNet50 as a benchmark, we were able to assess the efficacy of our proposed approaches in improving our task.

\end{itemize}

\subsection{Data Fusion} 
\begin{figure*}[h!t]
\centerline{\includegraphics[width=.7\linewidth]{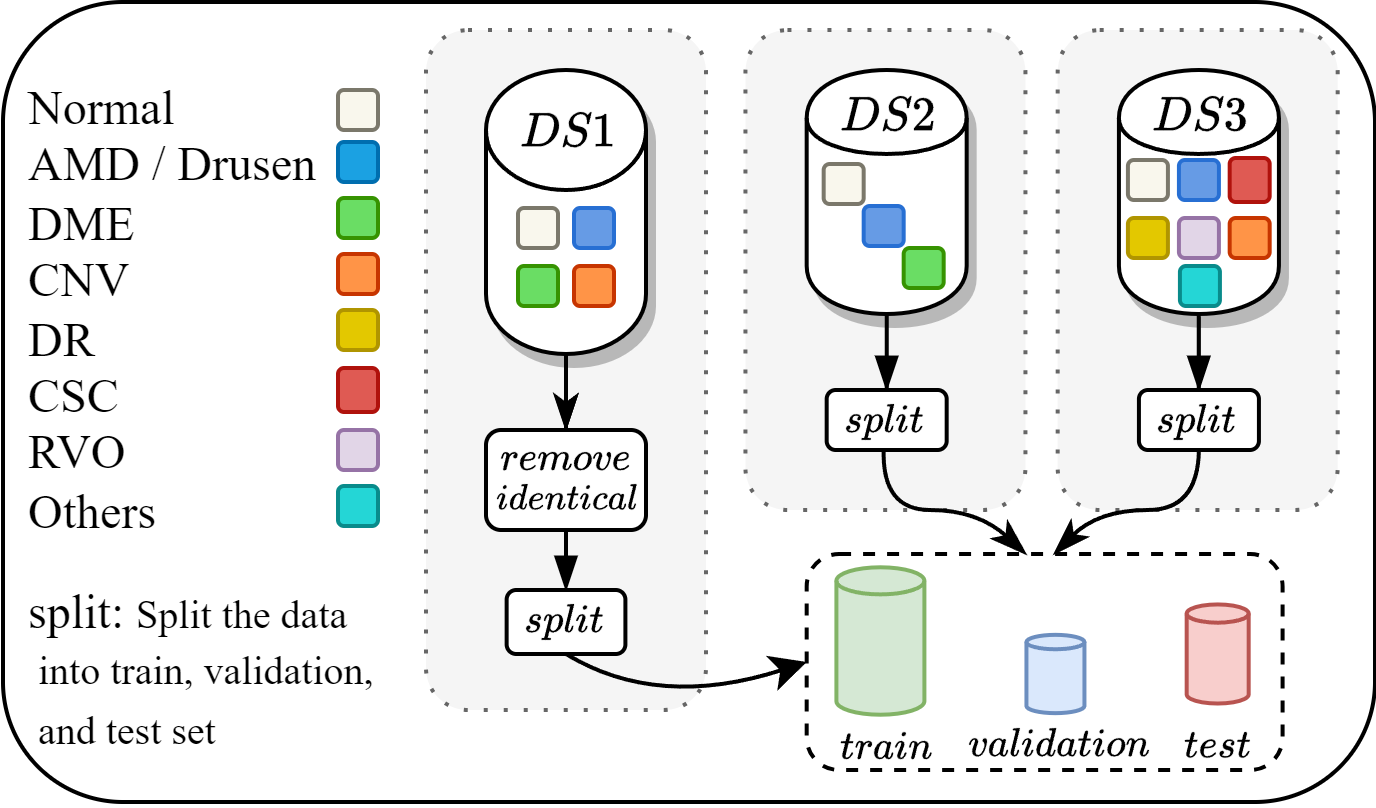}}
\caption{Illustration of the data combination process for the Self-Supervised Pre-training phase. Training and validation sets from three distinct Optical Coherence Tomography (OCT) datasets are merged to form a unified training and validation set, enhancing diversity and richness in the model's representation learning.
}
\label{fig-data-split}
\end{figure*}

For our study, we utilized three distinct datasets comprising Optical Coherence Tomography (OCT) images depicting various retinal diseases. Each dataset was partitioned into training, validation, and test sets. During the Self-Supervised Pre-training phase, we combined the training and validation sets from all three datasets into a unified training and validation set. This combination of data modalities aims to enhance the diversity and richness of the training data, facilitating a broader representation learning of the model. By training on this combined dataset, the model acquired a more comprehensive representation of OCT images, which ultimately contributed to improving the generalization of unseen data. Subsequently, in the Supervised Fine-tuning stage, the classifier underwent fine-tuning on each training set and was evaluated on all test sets to assess classification performance and generalization capabilities across different datasets. In Figure \ref{fig-data-split} the overall data combination process is shown.

\subsection{Self-Supervised Pre-training}
Self-supervised learning (SSL) is a method where models learn from unlabeled data by understanding its structure. In this study, we used a technique called Masked Autoencoder (MAE), which masks parts of input data randomly and trains the model to recreate the original by learning the representation of the input data. The MAE consists of two parts: an encoder and a decoder. We resized images to (224×224) and fed them through the encoder, which randomly masks 70\% of the input image. For the encoder component, we explored the performance of two distinct networks—Swin and SwinV2—as backbone architectures, facilitating a comprehensive investigation into their effectiveness.

\subsubsection{Swin-based MAE}
The encoder used in the Swin transformer-based Masked Autoencoder (MAE) is built with a Swin transformer backbone and has an embedding size of 96. The architecture of the Swin transformer has different numbers of layers at each stage (2, 2, 18, 2), which corresponds to the distribution of layers at each stage. The model employs shifted window attention mechanisms in each step to concentrate on local information within 4x4 patches, gradually constructing a global understanding through the connecting of shifted windows. With each step, the number of attention heads—which is set to (6, 12, 24, 48)—doubles, allowing the model to attend to progressively finer details and identify hierarchical features in the input image. In the meantime, a more expressive representation is made possible during the decoding process by the decoder, which is constructed with an embedding size of 768. 

The decoder network has a similar number of attention heads and layers as the encoder, along with Swin transformer layers that are set up to restore the spatial dimensions of the encoded features and a patch-expanding mechanism. This layer-wise design ensures a gradual reconstruction of the original image dimensions, facilitating effective decoding of the encoded representation acquired by the encoder.

\subsubsection{SwinV2-based MAE}
For this task, we utilize a SwinV2-based Masked Autoencoder (MAE), capitalizing on the superior performance of the SwinV2 network. While retaining the Swin-based decoder, we used the SwinV2 for the encoder component to address challenges related to training stability, high-resolution processing, and data efficiency. The enhanced capabilities of SwinV2 align with our requirements, creating a balance between detail-oriented feature extraction and computational efficiency. Leveraging an embedding dimension of 96, depths configured as (2, 2, 6, 2), and attention heads ranging from (3, 6, 12, 24), this customized approach allowed the SwinV2-based MAEs to excel in capturing intricate details essential for our task.

For another experiment, we enhanced the Swin-V2 encoder to increase its dimensionality and depth, resulting in a more complex network, which we denote as SwinV2-large. In this architecture, we adjusted the embedding dimension to 196, while configuring depths as (4, 4, 4, 4), and attention heads as(6, 12, 24, 48).

\subsection{Supervised Fine-Tuning}
We added a classification head in place of the decoder in the classifier network. The classification head employed a linear layer to process the encoder's features and produce class logits, which were then used for classification.

The linear layer consisted of three consecutive dense layers, each incorporating Rectified Linear Unit (ReLU) activation functions.  The input image was gradually transformed into highly encoded features by these layers. The initial layer had an input size equal to the dimension of the positional embedding from the encoder, with an output size of 512. The subsequent layer refined these features, mapping them to a 256-dimensional space, followed by a final layer compressing them into a 128-dimensional feature vector.

The purpose of this hierarchical transformation was to prime the model for successful classification tasks by highlighting and reducing the amount of intrinsic discriminative features in the input. During training, the linear layer learns weights and then class logits are transformed into class probabilities using the softmax function, allowing the model to predict classes accurately.

This methodology included fine-tuning one dataset, followed by assessing the model's classification performance on the corresponding test set. Additionally, two separate test sets from distinct datasets were utilized to assess the model's generalization and robustness. This iterative cross-data evaluation process was replicated across all three datasets, providing a thorough examination of the model's adaptability to varying data sources.

The development of Multi-OCT-SelfNet, which combines fusion data with supervised fine-tuning and self-supervised pre-training techniques, offers a strong framework for the classification of retinal diseases in Optical Coherence Tomography (OCT) images. This framework attempts to enhance the model's ability to generalize to new, unseen data by leveraging learned representations by combining the Masked Autoencoder (MAE) architecture with a subsequent classifier model. This holistic approach underscores the importance of utilizing fusion data with both self-supervised and supervised techniques to attain comprehensive and effective disease classification in OCT imaging.

The overall framework is given in Figure \ref{fig-framework} where it is shown the two-phase training approach, Firstly, it utilizes a masked image autoencoder for self-supervised learning from unlabeled images. Then, in the second phase, the pre-trained encoder is combined with a linear classifier for classification tasks, transferring the learned weights from the initial phase. This strategy optimizes the model's efficiency and effectiveness in handling classification tasks.

\subsection{Baseline Model}

ResNet50 stands as a versatile solution for handling intricate tasks like classifying age-related macular degeneration (AMD) and diabetic retinopathy using Optical Coherence Tomography (OCT) images, showcasing its efficacy in medical imaging analysis. To benchmark our proposed method's performance, we employed ResNet50 as our baseline model. The ResNet50 architecture comprises a 7×7 kernel convolution and a max pooling layer, succeeded by a series of convolutional layers with varying sizes and numbers of kernels. With 50 convolutional layers, the network is then followed by average pooling and fully connected layers, with the number of nodes matching the classes for multi-class classification, employing softmax activation.

\subsection{Loss Function}
For the pre-training stage, we have used a loss function, which only takes into account the pixels where the mask is active, and uses the mean squared error (MSE) between the predicted image and the original image. In Equation \ref{eq_loss_func}, the loss function is provided.
\begin{equation}\label{eq_loss_func}
\frac{1}{{m_{ratio}}} \times \frac{1}{N} \sum_{i=1}^{N} (\bar{x_i} - {x_i})^2 \times {m_i}
\end{equation}

Here $\bar{x_{i}}$ is the predicted image, ${x_{i}}$ is the original input image, ${m_{i}}$ is the mask, ${m_{ratio}}$ is the mask ratio and N is the number of total sample.

The MSE is multiplied by the mask to calculate the loss on the pixels where the mask is active. The mask ratio indicates the percentage of the image that is masked. Since the mask is being used to focus only on specific areas of the image, the loss is calculated by dividing the mean square error (MSE) by the mask ratio. This allows us to properly normalize the loss to the proportion of the image that is masked and scale the loss accordingly.

\subsection{Datasets}
\begin{figure*}[h!]
\centerline{\includegraphics[width=.8\linewidth]{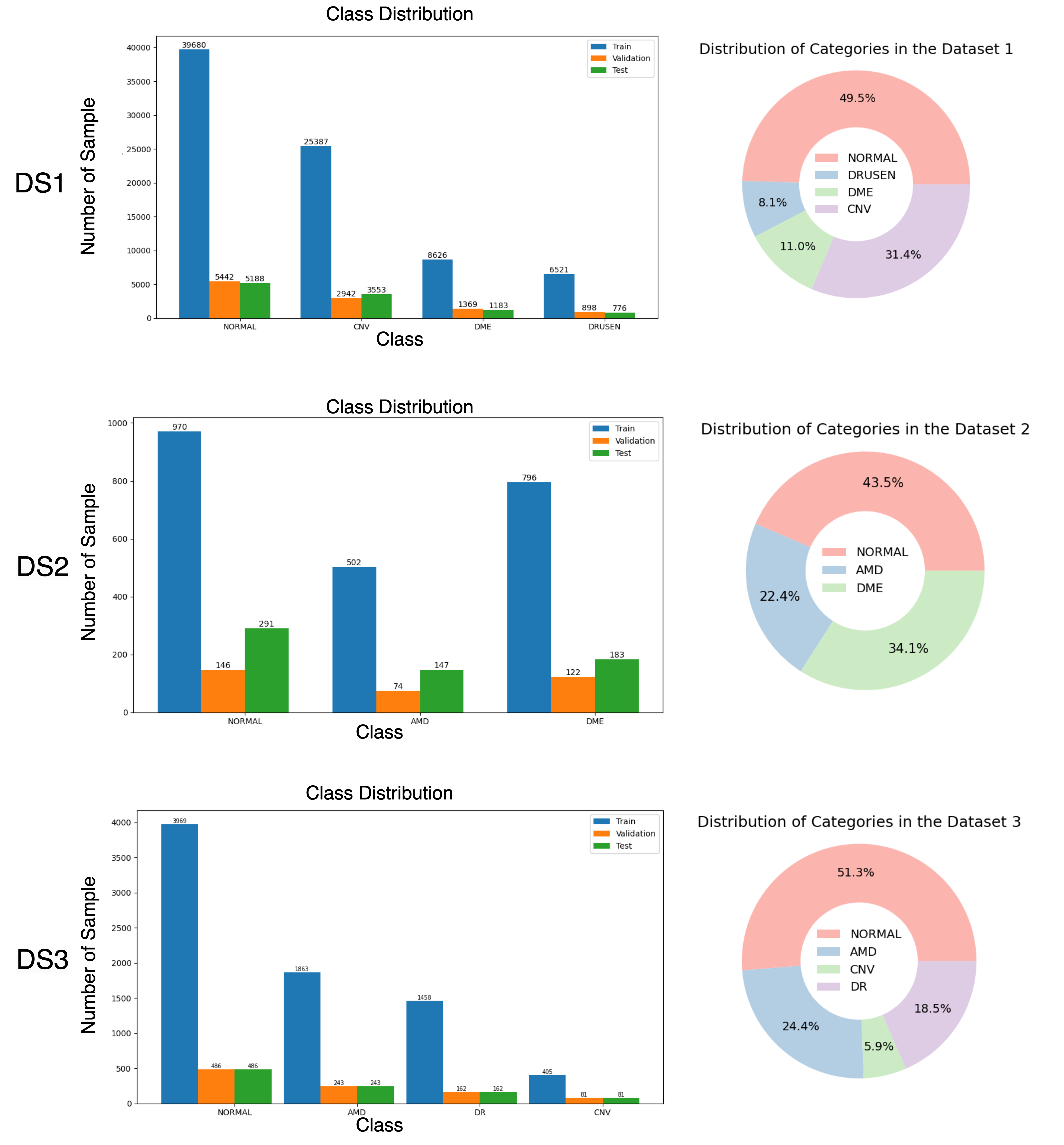}}
\caption{Distribution of Retinal Disease Samples Across Three Datasets: Grouped-bar diagrams show sample counts in training, validation, and test sets for each retinal disease category in datasets DS1, DS2, and DS3. The Donut charts display the overall percentage distribution per dataset.
}
\label{fig-dataset}
\end{figure*}

% \begin{figure*}[h!t]
% \centerline{\includegraphics[width=.8\linewidth]{acmlarge/figures/class-dis-across-datasets.png}}
% \caption{Heat map illustrating the distribution of classes across datasets. Each cell represents the count of a class within a dataset, allowing for the visualization and understanding of common classes shared among all datasets and variations in class representation.
% }
% \label{fig-data-heatmap}
% \end{figure*}
\subsubsection{DS1}

This dataset comprises a total of 109,559 Optical Coherence Tomography (OCT) retinal images acquired using Spectralis OCT from Heidelberg Engineering, Germany. These images are categorized into four classes: Normal, Choroidal Neovascularization (CNV), Diabetic Macular Edema (DME), and Drusen. Upon identifying identical images, as documented previously \cite{sinagh}, we followed their methodology to cleanse the dataset, resulting in 101,565 images. Subsequently, we reclassified Drusen images as Age-related Macular Degeneration (AMD). DS1 was then partitioned into training, testing, and validation sets using an 80\%, 10\%, and 10\% ratio, respectively. The distribution of samples across each category within each set is illustrated in Figure \ref{fig-dataset}.

\subsubsection{DS2}
The DS2 dataset is acquired using Heidelberg Engineering Spectralis SD-OCT protocols approved by the Institutional Review Board (IRB) \cite{srinivasan}. It consists of retinal images sourced from 45 subjects, distributed as follows: 15 normal subjects, 15 patients diagnosed with dry age-related macular degeneration (AMD), and 15 patients with diabetic macular edema (DME). To organize the dataset for training, validation, and testing purposes, we split the subjects accordingly: the first 10 subjects from each category were allocated to the training set, subjects 11 to 12 were assigned to the validation set, and subjects 13 to 15 were designated to the test set. Figure \ref{fig-dataset} shows the distribution of samples within each set, across each category.

\subsubsection{DS3}
This dataset comprising OCT images from 500 subjects, was obtained using two different fields of view: 3-mm and 6-mm. Each 3-mm file contains 304 scans per patient, while a 6-mm file contains 400 scans. Our analysis focused on the slice images of the fovea (image numbers 160-240 from the 6-mm scans), capturing the most prominent retinal features, while peripheral retinal sections were deemed of limited significance for classification. This dataset comprises categories such as NORMAL, AMD, CNV, DR, OTHERS, RVO, and CSC. Given the relatively low number of samples in the RVO and CSC categories, we excluded them. The "OTHERS" category comprises diseases, including retinal detachment (RD), retinal hemorrhage (RH), and retinitis pigmentosa (RP), among others, which were also discarded due to their lack of particularity. All OCT images were captured using a spectral-domain OCT system with a center wavelength of 840 nm (RTVue-XR, Optovue, CA) [ ]. Figure \ref{fig-dataset} provides an overview of the sample distribution across each category within each set.

\vspace{8pt}

In Figure \ref{fig-dataset} the distribution of retinal disease samples across training, validation, and test sets for each category in three datasets (DS1, DS2, and DS3) is shown. Grouped-bar diagrams depict the counts of samples in each category for training, validation, and test sets. Additionally, the donut charts illustrate the overall percentage distribution of samples for each dataset, providing a comprehensive view of the distribution of retinal disease samples in the study. Class presence and sample counts differ significantly; not all classes are present in all datasets in the same way. For example, although 'NORMAL' is present in all datasets, its frequency varies greatly: DS1 has a high count of 50,310 samples, DS3 has 4,941 samples, and DS2 has a significantly lower count of 1,407 samples. Similarly, the 'AMD' class exhibits varying degrees of representation, with 8,195 samples in DS1, 723 samples in DS2, and 2,349 samples in DS3. Such disparities extend further, with 'DME' being present in DS1 and DS2 but absent in DS3, while 'DR' finds exclusivity in DS3. These variations in class presence and sample counts highlight the complexities present in dataset composition and the implications for model development and evaluation.

% The DS1, DS2, and DS3 datasets are used to create the heatmap Figure \ref{fig-data-heatmap}, which shows the complex dynamics of class distribution throughout these datasets. The count of each class in each of the datasets is displayed in each cell of this heatmap. 

The model was fully trained with all classes found in datasets DS1, DS2, and DS3 during the self-supervised pre-training phase, which enabled it to fully understand the complexities of representation learning. Because of this inclusive training strategy, the network was able to capture a wide range of features and patterns that were present in the various classes. However, in the supervised fine-tuning stage, our focus narrowed to the classification tasks specifically targeting 'NORMAL' categories from the 'AMD', 'DME', 'CNV', and 'DR' classes. 
This two-stage training process, which includes thorough pre-training and specialized fine-tuning, strategically directs the model's learning of its representation and maximizes its performance for the desired classification goal. This methodology helps the model to be optimally tuned to achieve the intended classification objectives by shifting from a general comprehension of all classes to a targeted refinement of the desired classification tasks.

\section{Experiments}

\subsection{ Implementation Details}

\subsubsection{Self-Supervised Pre-training Implementation Details}
For this stage of self-supervised pre-training,  all experiments were conducted using the computational power of the NVIDIA Tesla V100 graphical processing unit (GPU).
Specific hyperparameters were selected to optimize model performance. The learning rate was set to $1.5\times10^{-4}$, and the Adam optimizer was employed with a weight decay of $0.05$ \cite{adamw}. Additionally, the optimizer utilized $\beta_1$ and $\beta_2$ values of 0.9 and 0.95, respectively. Input data consisted of batches comprising 32 normalized images. Training proceeded for a total of 100 epochs. To ensure the most robust configuration, the model with the lowest validation loss was saved for subsequent fine-tuning iterations.

\subsubsection{Supervised Fine-Tuning Implementation Details}

During the fine-tuning stage, similar to the training phase, experiments were conducted using the NVIDIA Tesla V100 GPU, and the hyperparameters were selected as follows, the learning rate was adjusted to $3\times10^{-4}$, and the Adam optimizer was employed with a weight decay of $1\times10^{-6}$. The optimizer utilized $\beta_1$ and $\beta_2$ values of 0.9 and 0.99, respectively. As the loss function, the categorical cross-entropy loss is employed. The training process involved the application of multiple data augmentation techniques, such as random resized crop, random horizontal flip, color jitter, random grayscale, and ImageNet normalization, to improve the robustness of the model. Training extended over 100 epochs, with early stopping criteria implemented using patience of 10 epochs. The model with the highest validation accuracy was saved for subsequent testing. 

During training and fine-tuning, all images were resized to 224 x 224. Python 3.10.9, Pytorch 1.12.1, and CUDA 11.2 were used to implement the codes.

\subsubsection{Baseline Model: ResNet-50 Implementation Details}

During the experiments with the baseline model, an NVIDIA GeForce RTX 3060 Ti GPU was used. The learning rate was set to $3\times10^{-4}$, accompanied by a weight decay of $10^{-6}$, a batch size of 24, and the utilization of the Adam optimizer with decoupled weight decay \cite{adamw}. Momentum and adaptive learning rate scaling were set by $\beta_1$ and $\beta_2$ values of 0.9 and 0.999, respectively. Training proceeded for a maximum of 100 epochs, with early stopping criteria based on validation loss, incorporating a patience of 10 epochs. To enhance the model's robustness and generalization capabilities, various data augmentation techniques were employed, including random rotation, horizontal flip, color jittering, Gaussian blurring, and elastic transform.

\subsection{Evaluation Metrics}

\red{As illustrated in Figure \ref{fig-dataset}, all datasets exhibit a substantial class imbalance, with a predominance of 'NORMAL' cases. Given this, relying solely on accuracy can be misleading. To mitigate this issue, we employ AUC-ROC as the primary evaluation measure, as it effectively provides a more robust evaluation by considering the trade-off between true positive and false positive rates across different thresholds.  Additionally, we incorporate accuracy, AUC-PR, and F1-score to offer a comprehensive evaluation.}

\subsubsection{Accuracy} The number of accurate predictions made by the model is known as accuracy, and it is determined by dividing the total number of predictions by the number of correct predictions. The accuracy formula is provided by Equation \ref{eq_accuracy}. 

\begin{equation}\label{eq_accuracy}
Accuracy = \frac{TP+TN}{TP+TN+FP+FN}
\end{equation}

Here, TP = True Positives, TN = True Negatives, FP = False Positives, and FN = False Negatives.

\subsubsection{AUC-ROC} One important metric for assessing the performance of the classifier is the Area Under the Receiver Operating Characteristic curve (AUC-ROC). Plotting each class's true positive rate against its false positive rate across a range of threshold values creates the ROC curve. The area under this curve, or AUC-ROC, gives a thorough overview of the model's performance in all classes and threshold settings. A higher AUC-ROC signifies better discrimination ability among the different classes, indicating superior classifier performance.
\subsubsection{AUC-PR} The Area Under the Precision-Recall curve (AUC-PR), a performance metric for classifier evaluation, is comparable to AUC-ROC. By plotting precision against recall across different threshold values, the PR curve emphasizes the trade-off between recall and precision. The AUC-PR quantifies the overall performance of the model across various threshold settings by measuring the area under this curve. A higher AUC-PR score indicates better classifier performance.

\subsubsection{F1-Score} Another performance evaluation metric that accounts for both recall and precision is the F1-Score. This metric is particularly helpful when dealing with data imbalances. By incorporating both precision and recall, the F1-Score provides a comprehensive evaluation of a classifier's effectiveness. This metric is especially suitable in situations where accurately capturing both positive and negative instances is critical for decision-making and model evaluation.

Equation \ref{eq_f1score} is used to calculate the F1-Score.
\begin{equation}\label{eq_f1score}
F1Score = \frac{2*Precision*Recall}{Precision + Recall}
\end{equation}

\subsubsection{Penalty-Based Performance Index}
To provide a quantitative assessment of the model's generalization performance, we created the Penalty-Based Performance Index to evaluate the models' performance across all test sets. This method calculates a penalty for each score by subtracting it from 1, representing the error rate. The scores considered for this calculation include accuracy, AUC-ROC, F1-score, and AUC-PR, encompassing various aspects of model performance. The average penalty for each model is then computed, indicating the overall error tendency which represents, on average, how much the model's score deviates from perfect score across all test sets. Finally, these average penalties are transformed into a scale of 1 to 100 to obtain a performance score for each model. A lower score indicates better performance, which means the model's accuracy is closer to a perfect score with minimal variance across the test sets.

The formula for calculating the Penalty-Based Performance Index for Model 1, Model 2, ..., Model $n$ with accuracy scores $A_1, A_2, \ldots, A_n$ is as follows:

\begin{equation}\label{eq_PerformanceScore}
Performance Index_x = \frac{\sum_{i=1}^{n} (1 - A_i)}{n} \times 100
\end{equation}

This evaluation method provides a quantitative measure of model performance, considering the accuracy and error rate, thus offering insights into the overall effectiveness of the models under assessment.

\subsection{Self-Supervised Pre-training Result}

\begin{figure*}[h!t]
\centerline{\includegraphics[width=.98\linewidth]{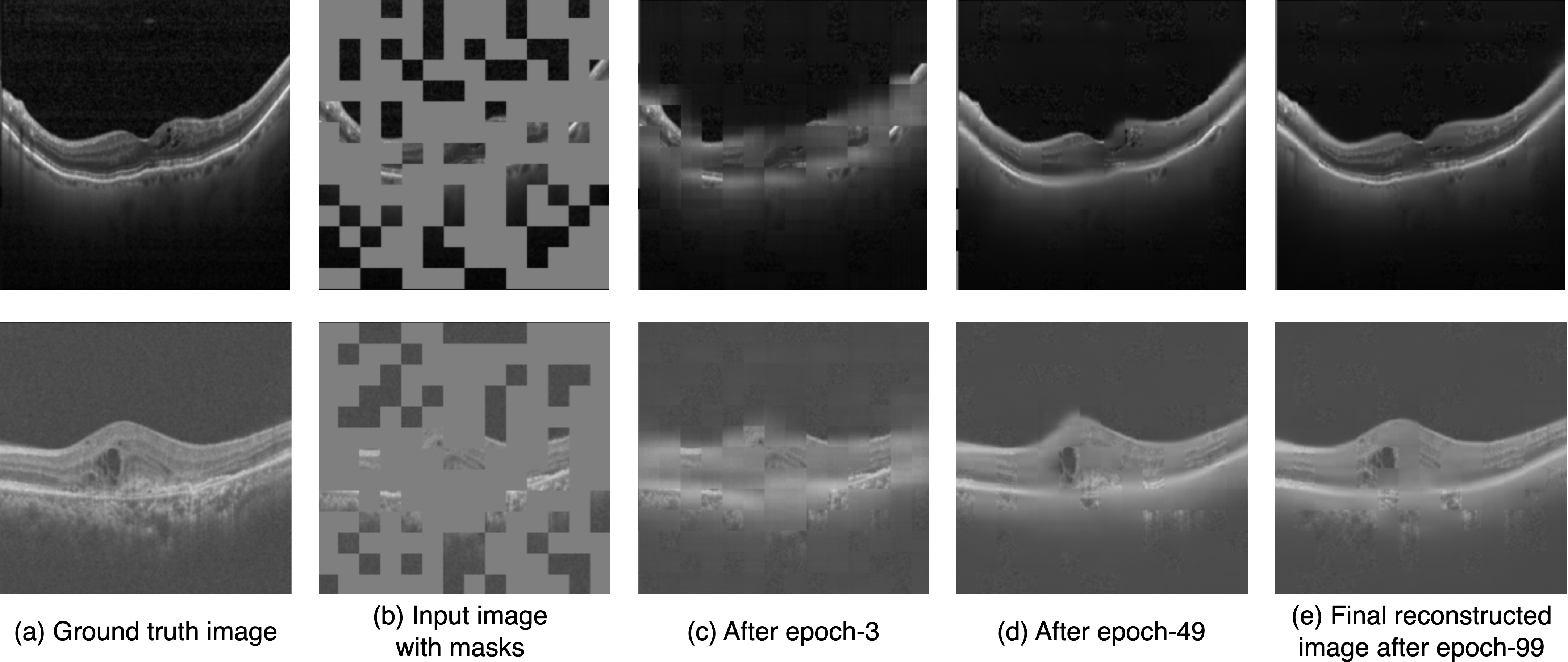}}
\caption{The progression results of the Multi-OCT-SelfNet-SwinV2 model on sample validation images across various epochs illustrate its learning process in reconstructing input images. From left to right, (a) is the corresponding ground truth image, (b) is the masked image input, and (c)-(e) is the reconstructed images in different epochs.
}
\label{fig-ssl-result}
\end{figure*}

\begin{figure*}[h!t]
\centerline{\includegraphics[width=.75\linewidth]{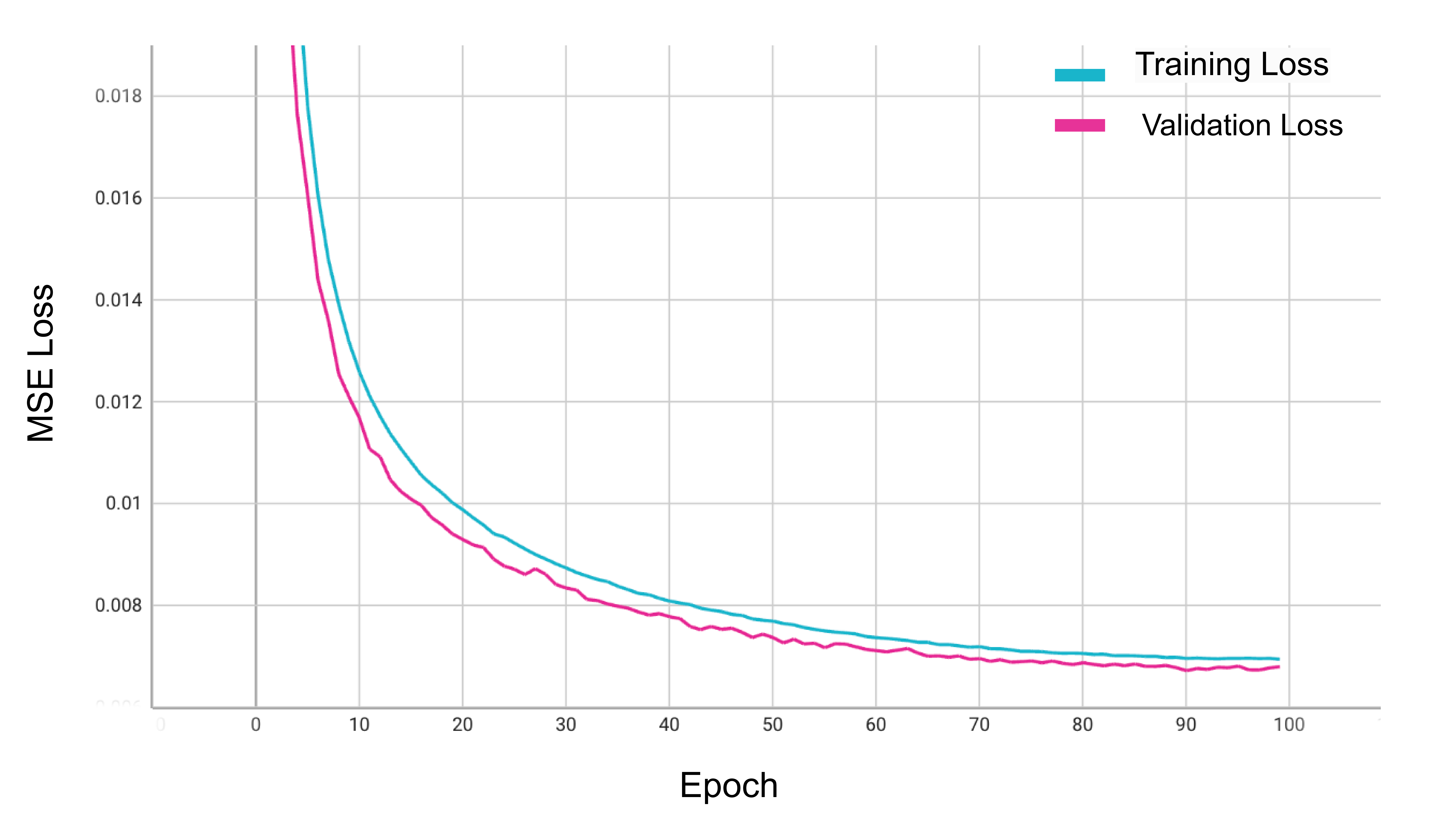}}
\caption{The training and validation MSE Loss curves of Multi-OCT-SelfNet with
SwinV2 backbone, trained with the combined dataset.
}
\label{fig-ssl-loss}
\end{figure*}

In our ablation study, we conducted pre-training for 100 epochs on three transformer-based networks. One network utilized the Swin architecture as its backbone, while the other two employed distinct variations of SwinV2 as backbones—referred to as SwinV2 and SwinV2-large, respectively. The objective was to assess the efficacy of these models and their performance in subsequent tasks. In particular, after 100 training epochs, the SwinV2-based MAE demonstrated remarkable proficiency, obtaining a mean squared error (MSE) loss of 0.007  as shown in Fig \ref{fig-ssl-loss}. Moreover, Fig \ref{fig-ssl-result} provides a detailed visualization of the masked image input and the corresponding image reconstruction from the SwinV2-based MAE at various epochs, offering insights into the model's learning dynamics throughout the training process. While the reconstructed images may display imperfections, our project prioritized the capture of intricate image structures and patterns over flawless reconstructions. Consequently, in subsequent tasks, we employed these pre-trained weights, capitalizing on their learned representations, rather than initializing the classifiers with random weights. By applying the learned knowledge encoded in the pre-trained weights, this method allowed us to take advantage of the benefits of the transfer learning approach, facilitating enhanced performance and accelerated convergence in downstream tasks.

\subsection{Supervised Fine-tuning Result}

\subsubsection{Performance Comparison with Different Encoder Network}
We evaluated different encoder networks against the baseline ResNet-50 model in our extensive experiment. To ensure consistency, we utilized the same encoder from the SSL networks and employed transfer learning to transfer the learned weights. For downstream classification tasks, we also integrated a classifier network. Each supervised network underwent fine-tuning individually for every dataset, followed by evaluations on both the respective test set and the test sets from other datasets. This comprehensive evaluation allowed us to observe the network's performance on previously unseen test data and to observe its generalization capability. The performance of each model was then compared with that of the baseline model ResNet-50, which underwent training on each dataset and subsequent evaluation on all three test sets. Throughout this experimental process, we applied data augmentation techniques to enhance robustness. Performance metrics including Accuracy, AUC-ROC, AUC-PR, and F1-Score were employed to measure effectiveness. Analysis of Table \ref{tab:result-encoder} revealed the consistent performance of the self-supervised fine-tuning approach over the baseline model, with the SwinV2-based classifier demonstrating the most reliable performance across all test sets, particularly in scenarios with smaller datasets such as Dataset-2 and Dataset-3. While Dataset-1, with its larger sample size, yielded similar performance between the proposed framework and the baseline model, it's noteworthy that our method showcased superior performance on smaller datasets as well. The generalization capabilities during off-domain evaluation,  when the model was fine-tuned on one dataset and evaluated on other test sets, significant performance boosts were observed in our proposed method. For instance, while fine-tuning Dataset-1, our Multi-OCT-SelfNet-SwinV2 model achieved AUC-ROC scores of 0.79, 0.97, and 0.90, along with AUC-PR of 0.58, 0.94, and 0.70 on Test Set-1, Test Set-2, and Test Set-3, respectively. In contrast, the baseline model attained AUC-ROC scores of 0.65, 0.98, and 0.59, along with AUC-PR of 0.39, 0.95, and 0.57 on the respective test sets. These results underscore the superior generalization capability of our proposed method, highlighting its potential for robust performance across diverse datasets. 

\begin{figure*}[h!t]
\centerline{\includegraphics[width=.75\linewidth]{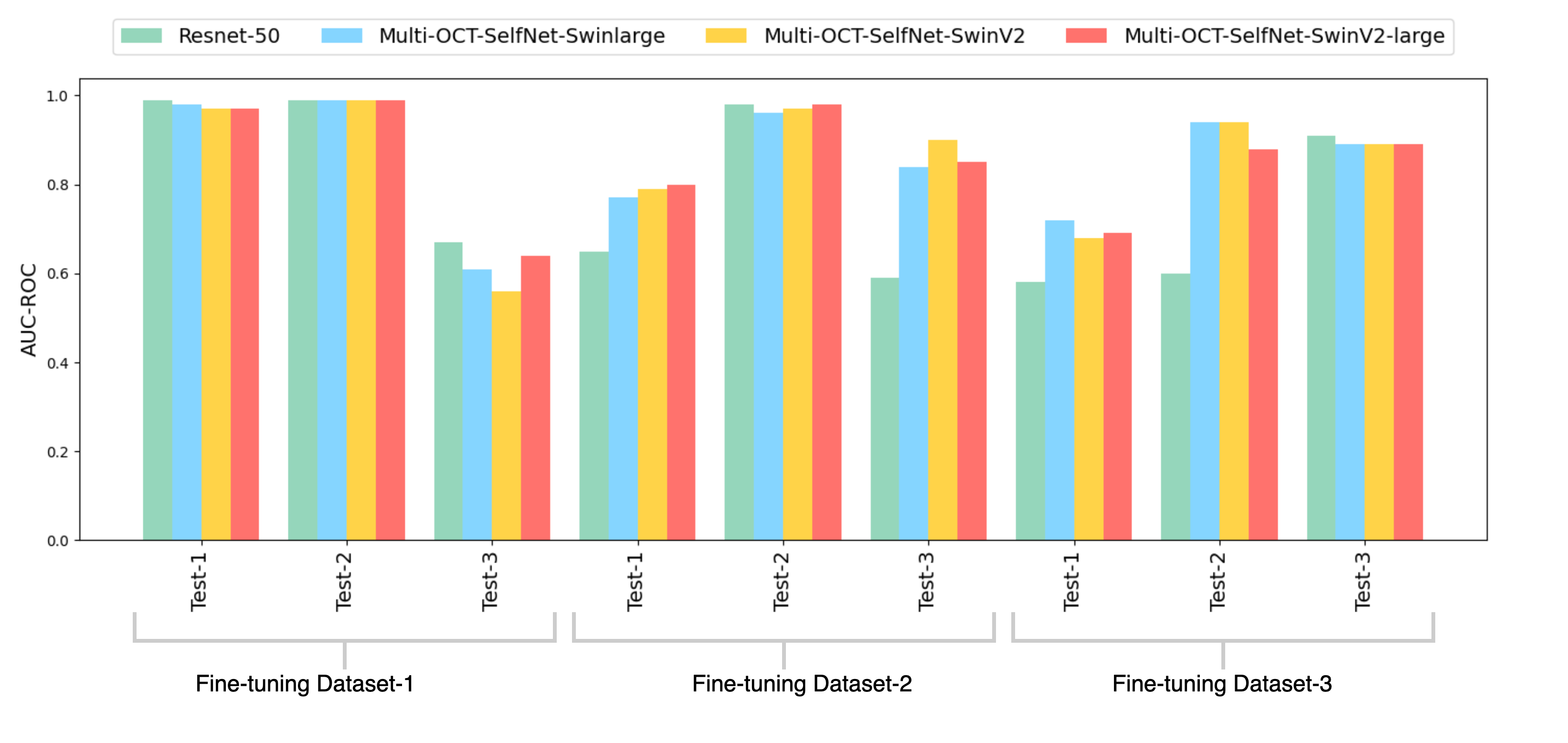}}
\caption{Comparison of AUC-ROC scores for different classifiers across three datasets and test sets. Each group represents AUC-ROC scores for classifiers within a specific dataset and test set combination, highlighting performance variations.
}
\label{fig-table-bar-1}
\end{figure*}

Figure \ref{fig-table-bar-1} presents a bar chart comparing AUC-ROC scores for various classifiers across three datasets and test sets. The grouped bars facilitate a direct performance comparison, revealing that Multi-OCT-SelfNet generally surpasses the baseline ResNet-50. This chart highlights the performance of each classifier across different datasets and underscores Multi-OCT-SelfNet's superior AUC-ROC scores. AUC-ROC is an ideal metric for evaluating classifiers on imbalanced binary datasets. Therefore, AUC-ROC was chosen for our bar chart to deliver a comprehensive assessment of classifier performance across the datasets and test sets.

To further evaluate generalizability and domain adaptability across other datasets, we employed a performance metric known as the Penalty-Based Performance Index, as presented in Table \ref{tab:p-index}. This index aggregates scores from three distinct test sets and calculates penalty scores for each specific metric. A lower penalty score indicates superior generalization performance, highlighting the model's ability to adapt across diverse datasets. Upon observing the results in the table, it becomes evident that the proposed framework consistently exhibits better generalization performance across various metrics. This reaffirms the robustness and adaptability of our model.

% Please add the following required packages to your document preamble:
% \usepackage{multirow}
% \usepackage[table,xcdraw]{xcolor}
% Beamer presentation requires \usepackage{colortbl} instead of \usepackage[table,xcdraw]{xcolor}
\begin{table}[h]
\centering
\small
\scriptsize % Reduce font size
\caption{Analyzing the Performance with Different Encoder Networks: Comparison of our framework with baseline methods (ResNet-50) on the test sets from three datasets, evaluating multi-class classification performance in terms of accuracy, AUC-ROC, AUC-PR, and F1-score.}
\label{tab:result-encoder}
\begin{tabular}{|l|l|lll|lll|lll|lll|}
\hline
 &
   &
  \multicolumn{3}{c|}{AUC-ROC} &
  \multicolumn{3}{c|}{Accuracy} &
  \multicolumn{3}{c|}{AUC-PR} &
  \multicolumn{3}{c|}{F1-Score} \\ \cline{3-14} 
\multirow{-2}{*}{Dataset} &
  \multirow{-2}{*}{Classifier Name} &
  \multicolumn{1}{l|}{Test-1} &
  \multicolumn{1}{l|}{Test-2} &
  Test-3 &
  \multicolumn{1}{l|}{Test-1} &
  \multicolumn{1}{l|}{Test-2} &
  Test-3 &
  \multicolumn{1}{l|}{Test-1} &
  \multicolumn{1}{l|}{Test-2} &
  Test-3 &
  \multicolumn{1}{l|}{Test-1} &
  \multicolumn{1}{l|}{Test-2} &
  Test-3 \\ \hline
 &
  \cellcolor[HTML]{F3F3F3}Resnet-50-Multi &
  
  \multicolumn{1}{l|}{\cellcolor[HTML]{F3F3F3}0.99} &
  \multicolumn{1}{l|}{\cellcolor[HTML]{F3F3F3}0.99} &
  \cellcolor[HTML]{F3F3F3}0.67 &
  \multicolumn{1}{l|}{\cellcolor[HTML]{F3F3F3}0.95} &
  \multicolumn{1}{l|}{\cellcolor[HTML]{F3F3F3}0.92} &
  \cellcolor[HTML]{F3F3F3}0.24 &
  \multicolumn{1}{l|}{\cellcolor[HTML]{F3F3F3}0.96} &
  \multicolumn{1}{l|}{\cellcolor[HTML]{F3F3F3}0.99} &
  \cellcolor[HTML]{F3F3F3}0.53 &
  \multicolumn{1}{l|}{\cellcolor[HTML]{F3F3F3}0.95} &
  \multicolumn{1}{l|}{\cellcolor[HTML]{F3F3F3}0.92} &
  \cellcolor[HTML]{F3F3F3}0.25 \\ \cline{2-14} 
 &
  \cellcolor[HTML]{F3F3F3}Multi-OCT-SelfNet-Swinlarge &

  \multicolumn{1}{l|}{\cellcolor[HTML]{F3F3F3}0.98} &
  \multicolumn{1}{l|}{\cellcolor[HTML]{F3F3F3}0.99} &
  \cellcolor[HTML]{F3F3F3}0.61 &
  \multicolumn{1}{l|}{\cellcolor[HTML]{F3F3F3}0.91} &
  \multicolumn{1}{l|}{\cellcolor[HTML]{F3F3F3}0.92} &
  \cellcolor[HTML]{F3F3F3}0.41 &
  \multicolumn{1}{l|}{\cellcolor[HTML]{F3F3F3}0.89} &
  \multicolumn{1}{l|}{\cellcolor[HTML]{F3F3F3}0.99} &
  \cellcolor[HTML]{F3F3F3}0.43 &
  \multicolumn{1}{l|}{\cellcolor[HTML]{F3F3F3}0.90} &
  \multicolumn{1}{l|}{\cellcolor[HTML]{F3F3F3}0.94} &
  \cellcolor[HTML]{F3F3F3}0.54 \\ \cline{2-14} 
 &
  \cellcolor[HTML]{F3F3F3}Multi-OCT-SelfNet-SwinV2 &
  
  \multicolumn{1}{l|}{\cellcolor[HTML]{F3F3F3}0.97} &
  \multicolumn{1}{l|}{\cellcolor[HTML]{F3F3F3}0.99} &
  \cellcolor[HTML]{F3F3F3}0.56 &
  \multicolumn{1}{l|}{\cellcolor[HTML]{F3F3F3}0.90} &
  \multicolumn{1}{l|}{\cellcolor[HTML]{F3F3F3}0.86} &
  \cellcolor[HTML]{F3F3F3}0.46 &
  \multicolumn{1}{l|}{\cellcolor[HTML]{F3F3F3}0.89} &
  \multicolumn{1}{l|}{\cellcolor[HTML]{F3F3F3}0.98} &
  \cellcolor[HTML]{F3F3F3}0.42 &
  \multicolumn{1}{l|}{\cellcolor[HTML]{F3F3F3}0.90} &
  \multicolumn{1}{l|}{\cellcolor[HTML]{F3F3F3}0.91} &
  \cellcolor[HTML]{F3F3F3}0.87 \\ \cline{2-14} 
\multirow{-4}{*}{DS-1} &
  \cellcolor[HTML]{F3F3F3}Multi-OCT-SelfNet-SwinV2-large &
  
  \multicolumn{1}{l|}{\cellcolor[HTML]{F3F3F3}0.97} &
  \multicolumn{1}{l|}{\cellcolor[HTML]{F3F3F3}0.99} &
  \cellcolor[HTML]{F3F3F3}0.64 &
  \multicolumn{1}{l|}{\cellcolor[HTML]{F3F3F3}0.91} &
  \multicolumn{1}{l|}{\cellcolor[HTML]{F3F3F3}0.91} &
  \cellcolor[HTML]{F3F3F3}0.40 &
  \multicolumn{1}{l|}{\cellcolor[HTML]{F3F3F3}0.90} &
  \multicolumn{1}{l|}{\cellcolor[HTML]{F3F3F3}0.99} &
  \cellcolor[HTML]{F3F3F3}0.44 &
  \multicolumn{1}{l|}{\cellcolor[HTML]{F3F3F3}0.91} &
  \multicolumn{1}{l|}{\cellcolor[HTML]{F3F3F3}0.93} &
  \cellcolor[HTML]{F3F3F3}0.56 \\ \hline
 &
  \cellcolor[HTML]{F3F3F3}Resnet-50-Multi &
  
  \multicolumn{1}{l|}{0.65} &
  \multicolumn{1}{l|}{0.98} &
  0.59 &
  \multicolumn{1}{l|}{0.29} &
  \multicolumn{1}{l|}{0.87} &
  0 &
  \multicolumn{1}{l|}{0.39} &
  \multicolumn{1}{l|}{0.95} &
  0.57 &
  \multicolumn{1}{l|}{0.31} &
  \multicolumn{1}{l|}{0.87} &
  0 \\ \cline{2-14} 
 &
  \cellcolor[HTML]{F3F3F3}Multi-OCT-SelfNet-Swinlarge &
  
  \multicolumn{1}{l|}{0.77} &
  \multicolumn{1}{l|}{0.96} &
  0.84 &
  \multicolumn{1}{l|}{0.62} &
  \multicolumn{1}{l|}{0.88} &
  0.45 &
  \multicolumn{1}{l|}{0.54} &
  \multicolumn{1}{l|}{0.92} &
  0.64 &
  \multicolumn{1}{l|}{0.65} &
  \multicolumn{1}{l|}{0.87} &
  0.59 \\ \cline{2-14} 
 &
  \cellcolor[HTML]{F3F3F3}Multi-OCT-SelfNet-SwinV2 &
 
  \multicolumn{1}{l|}{0.79} &
  \multicolumn{1}{l|}{0.97} &
  0.90 &
   \multicolumn{1}{l|}{0.65} &
  \multicolumn{1}{l|}{0.86} &
  0.45 &
  \multicolumn{1}{l|}{0.58} &
  \multicolumn{1}{l|}{0.94} &
  0.70 &
  \multicolumn{1}{l|}{0.68} &
  \multicolumn{1}{l|}{0.86} &
  0.54 \\ \cline{2-14} 
\multirow{-4}{*}{DS-2} &
  \cellcolor[HTML]{F3F3F3}Multi-OCT-SelfNet-SwinV2-large &
  
  \multicolumn{1}{l|}{0.80} &
  \multicolumn{1}{l|}{0.98} &
  0.85 &
  \multicolumn{1}{l|}{0.68} &
  \multicolumn{1}{l|}{0.93} &
  0.28 &
  \multicolumn{1}{l|}{0.61} &
  \multicolumn{1}{l|}{0.97} &
  0.63 &
  \multicolumn{1}{l|}{0.70} &
  \multicolumn{1}{l|}{0.93} &
  0.73 \\ \hline
 &
  \cellcolor[HTML]{F3F3F3}Resnet-50-Multi &
  
  \multicolumn{1}{l|}{\cellcolor[HTML]{F3F3F3}0.58} &
  \multicolumn{1}{l|}{\cellcolor[HTML]{F3F3F3}0.60} &
  \cellcolor[HTML]{F3F3F3}0.91 &
  \multicolumn{1}{l|}{\cellcolor[HTML]{F3F3F3}0.55} &
  \multicolumn{1}{l|}{\cellcolor[HTML]{F3F3F3}0.66} &
  \cellcolor[HTML]{F3F3F3}0.86 &
  \multicolumn{1}{l|}{\cellcolor[HTML]{F3F3F3}0.45} &
  \multicolumn{1}{l|}{\cellcolor[HTML]{F3F3F3}0.82} &
  \cellcolor[HTML]{F3F3F3}0.74 &
  \multicolumn{1}{l|}{\cellcolor[HTML]{F3F3F3}0.39} &
  \multicolumn{1}{l|}{\cellcolor[HTML]{F3F3F3}0.53} &
  \cellcolor[HTML]{F3F3F3}0.82 \\ \cline{2-14} 
 &
  \cellcolor[HTML]{F3F3F3}Multi-OCT-SelfNet-Swinlarge &

  \multicolumn{1}{l|}{\cellcolor[HTML]{F3F3F3}0.72} &
  \multicolumn{1}{l|}{\cellcolor[HTML]{F3F3F3}0.94} &
  \cellcolor[HTML]{F3F3F3}0.89 &
  \multicolumn{1}{l|}{\cellcolor[HTML]{F3F3F3}0.51} &
  \multicolumn{1}{l|}{\cellcolor[HTML]{F3F3F3}0.86} &
  \cellcolor[HTML]{F3F3F3}0.88 &
  \multicolumn{1}{l|}{\cellcolor[HTML]{F3F3F3}0.52} &
  \multicolumn{1}{l|}{\cellcolor[HTML]{F3F3F3}0.90} &
  \cellcolor[HTML]{F3F3F3}0.85 &
  \multicolumn{1}{l|}{\cellcolor[HTML]{F3F3F3}0.49} &
  \multicolumn{1}{l|}{\cellcolor[HTML]{F3F3F3}0.82} &
  \cellcolor[HTML]{F3F3F3}0.85 \\ \cline{2-14} 
 &
  \cellcolor[HTML]{F3F3F3}Multi-OCT-SelfNet-SwinV2 &
  
  \multicolumn{1}{l|}{\cellcolor[HTML]{F3F3F3}0.68} &
  \multicolumn{1}{l|}{\cellcolor[HTML]{F3F3F3}0.94} &
  \cellcolor[HTML]{F3F3F3}0.89 &
  \multicolumn{1}{l|}{\cellcolor[HTML]{F3F3F3}0.45} &
  \multicolumn{1}{l|}{\cellcolor[HTML]{F3F3F3}0.84} &
  \cellcolor[HTML]{F3F3F3}0.86 &
  \multicolumn{1}{l|}{\cellcolor[HTML]{F3F3F3}0.49} &
  \multicolumn{1}{l|}{\cellcolor[HTML]{F3F3F3}0.93} &
  \cellcolor[HTML]{F3F3F3}0.75 &
  \multicolumn{1}{l|}{\cellcolor[HTML]{F3F3F3}0.49} &
  \multicolumn{1}{l|}{\cellcolor[HTML]{F3F3F3}0.84} &
  \cellcolor[HTML]{F3F3F3}0.83 \\ \cline{2-14} 
\multirow{-4}{*}{DS-3} &
  \cellcolor[HTML]{F3F3F3}Multi-OCT-SelfNet-SwinV2-large &
  
  \multicolumn{1}{l|}{\cellcolor[HTML]{F3F3F3}0.69} &
  \multicolumn{1}{l|}{\cellcolor[HTML]{F3F3F3}0.88} &
  \cellcolor[HTML]{F3F3F3}0.89 &
  \multicolumn{1}{l|}{\cellcolor[HTML]{F3F3F3}0.52} &
  \multicolumn{1}{l|}{\cellcolor[HTML]{F3F3F3}0.79} &
  \cellcolor[HTML]{F3F3F3}0.87 &
  \multicolumn{1}{l|}{\cellcolor[HTML]{F3F3F3}0.49} &
  \multicolumn{1}{l|}{\cellcolor[HTML]{F3F3F3}0.85} &
  \cellcolor[HTML]{F3F3F3}0.79 &
  \multicolumn{1}{l|}{\cellcolor[HTML]{F3F3F3}0.51} &
  \multicolumn{1}{l|}{\cellcolor[HTML]{F3F3F3}0.71} &
  \cellcolor[HTML]{F3F3F3}0.85 \\ \hline
\end{tabular}
\end{table}

% Please add the following required packages to your document preamble:
% \usepackage{multirow}
% \usepackage[table,xcdraw]{xcolor}
% Beamer presentation requires \usepackage{colortbl} instead of \usepackage[table,xcdraw]{xcolor}
\begin{table}[h]
\centering
\scriptsize
\caption{Analyzing Model’s Generalization Performance on Different Encoder Networks: Comparison of penalty-based performance scores across three test sets for different models and three datasets.}
\label{tab:p-index}
\begin{tabular}{|l|l|l|l|l|l|}
\hline
 &
  \multicolumn{1}{c|}{} &
  \multicolumn{1}{c|}{} &
  \multicolumn{1}{c|}{} &
  \multicolumn{1}{c|}{} &
  \multicolumn{1}{c|}{} \\
\multirow{-2}{*}{Dataset} &
  \multicolumn{1}{c|}{\multirow{-2}{*}{Classifier Name}} &
  \multicolumn{1}{c|}{\multirow{-2}{*}{P-Index of AUC-ROC}} &
  \multicolumn{1}{c|}{\multirow{-2}{*}{P-Index of Accuracy}} &
  \multicolumn{1}{c|}{\multirow{-2}{*}{P-Index of AUC-PR}} &
  \multicolumn{1}{c|}{\multirow{-2}{*}{P-Index of F1-Score}} \\ \hline
\rowcolor[HTML]{F3F3F3} 
\cellcolor[HTML]{F3F3F3} &
  Resnet-50-Multi &
  \textbf{11.66 }&
  29.6 &
  
  \textbf{17.33 } &
  29.33 \\ \cline{2-6} 
\rowcolor[HTML]{F3F3F3} 
\cellcolor[HTML]{F3F3F3} &
  Multi-OCT-SelfNet-Swinlarge &
  14.00 &
  \textbf{25.33} &
  
  23.0 &
  20.66 \\ \cline{2-6} 
\rowcolor[HTML]{F3F3F3} 
\cellcolor[HTML]{F3F3F3} &
  Multi-OCT-SelfNet-SwinV2 &
  16.0 &
  26.0 &
  
  23.67 &
  \textbf{10.67} \\ \cline{2-6} 
\rowcolor[HTML]{F3F3F3} 
\multirow{-4}{*}{\cellcolor[HTML]{F3F3F3}Dataset-1} &
  Multi-OCT-SelfNet-SwinV2-large &
  
  13.33 &
  25.99 &
  22.33 &
  19.99 \\ \hline
 &
  \cellcolor[HTML]{F3F3F3}Resnet-50-Multi &
  
  26.0 &
  61.33 &
  36.33 &
  60.66 \\ \cline{2-6} 
 &
  \cellcolor[HTML]{F3F3F3}OCT-SelfNet-Swinlarge-Multi &
  
  14.33 &
  35.0 &
  30.0 &
  29.66 \\ \cline{2-6} 
 &
  \cellcolor[HTML]{F3F3F3}OCT-SelfNet-SwinV2-Multi &
  
  \textbf{11.33} &
  \textbf{34.67} &
  \textbf{26.00} &
  30.66 \\ \cline{2-6} 
\multirow{-4}{*}{Dataset-2} &
  \cellcolor[HTML]{F3F3F3}OCT-SelfNet-SwinV2-large-Multi &
  
  12.33 &
  36.99 &
  26.33 &
  \textbf{21.33} \\ \hline
\rowcolor[HTML]{F3F3F3} 
\cellcolor[HTML]{F3F3F3} &
  Resnet-50-Multi &
  
  30.33 &
  31.0 &
  33.0 &
  42.00 \\ \cline{2-6} 
\rowcolor[HTML]{F3F3F3} 
\cellcolor[HTML]{F3F3F3} &
  OCT-SelfNet-Swinlarge-Multi &
  
  \textbf{15.00} &
  \textbf{25.0} &
  \textbf{24.33} &
  \textbf{28.00} \\ \cline{2-6} 
\rowcolor[HTML]{F3F3F3} 
\cellcolor[HTML]{F3F3F3} &
  OCT-SelfNet-SwinV2-Multi &
  
  16.33 &
  28.33 &
  27.67 &
  \textbf{28.00} \\ \cline{2-6} 
\rowcolor[HTML]{F3F3F3} 
\multirow{-4}{*}{\cellcolor[HTML]{F3F3F3}Dataset-3} &
  OCT-SelfNet-SwinV2-large-Multi &
 
  18.00 &
  27.33 &
  28.99 &
  31.0 \\ \hline
\end{tabular}
\end{table}

\subsubsection{Performance Evaluation without Data Fusion during Pre-training Phase}

In this experiment, we assessed the performance without the data fusion during the SSL pre-training phase. We opted to pre-train the SSL model with each dataset individually and subsequently evaluated the classifier's performance solely on the respective test sets. This approach was taken to gain deeper insights into the contribution of data fusion to the classifier's performance. By restricting the training and evaluation processes to a single dataset, we aimed to investigate the impact of data fusion on the classifier's performance. This analysis provided valuable insights into the role of data integration of different modalities in enhancing classifier performance.

Table \ref{tab:result-wo-com-ds} presents the results of the on-domain evaluation for our Multi-OCT-SelfNet framework utilizing the SwinV2 backbone, excluding the data fusion step in the pre-training phase. This analysis aims to quantify the impact of data fusion on performance enhancement. Notably, for Dataset-2, the achieved accuracy, AUC-ROC, AUC-PR, and F1-Score were 0.74, 0.89, 0.80, and 0.74, respectively. Upon incorporating data fusion, substantial improvements were observed, with scores reaching 0.86, 0.97, 0.94, and 0.86, respectively, indicating a significant boost in performance which can be seen from previous Table \ref{tab:result-encoder}. Similarly, for Dataset-3, the scores stood at 0.58, 0.67, 0.35, and 0.47, while data fusion resulted in notable enhancements, yielding scores of 0.86, 0.89, 0.49, and 0.83, respectively, thus demonstrating consistent performance improvements. Conversely, for Dataset-1, where the dataset size was considerably larger and already facilitated robust training, the impact of data fusion on performance enhancement was relatively modest. Although there were slight improvements in accuracy, AUC-ROC, AUC-PR, and F1-Score, from 0.89, 0.97, 0.88, and 0.90 to 0.91, 0.97, 0.89, and 0.90, respectively, these enhancements were not as substantial. 

These findings highlight the significance of data fusion, particularly in scenarios with smaller datasets, where it significantly contributes to performance improvement.

% Please add the following required packages to your document preamble:
\begin{table}[]
\centering
\caption{Analyzing the Performance without Data Fusion during Pre-training Phase: Comparison of SwinV2-based classifier performance, where the encoder is pre-trained on individual training sets and the classifier is subsequently fine-tuned on the same training set, followed by evaluation on respective test sets.}
\label{tab:result-wo-com-ds}
\begin{tabular}{|l|l|l|l|l|l|}
\hline
Dataset Name & Classifier Name          & AUC-ROC  & Accuracy & AUC-PR & F1-Score \\ \hline
Dataset-1    & Multi-OCT-SelfNet-SwinV2      &   0.97  &     0.89    & 0.88       &  0.90        \\ \hline
Dataset-2    & Multi-OCT-SelfNet-SwinV2      & 0.89  & 0.74  & 0.80   & 0.74     \\ \hline
Dataset-3    & Multi-OCT-SelfNet-SwinV2     & 0.67  & 0.58   & 0.35   & 0.47     \\ \hline
\end{tabular}
\end{table}

\subsubsection{Performance Evaluation on the Effect of Self-supervised Pre-training}
In this ablation study, we investigated the impact of self-supervised pre-training on classifier performance in two key aspects: accurate label classification and generalization to diverse test sets. Instead of fine-tuning a pre-trained model as in the previous experiment, we trained the model from scratch. We used random initialization of the model’s parameters while training to evaluate the model's performance. We found that, even without fine-tuning, the classifier achieved comparable accuracy to the fine-tuned model when evaluated on the same dataset and test sets. However,  the model required a higher number of epochs to converge.

\begin{figure*}[h!]
\centerline{\includegraphics[width=.75\linewidth]{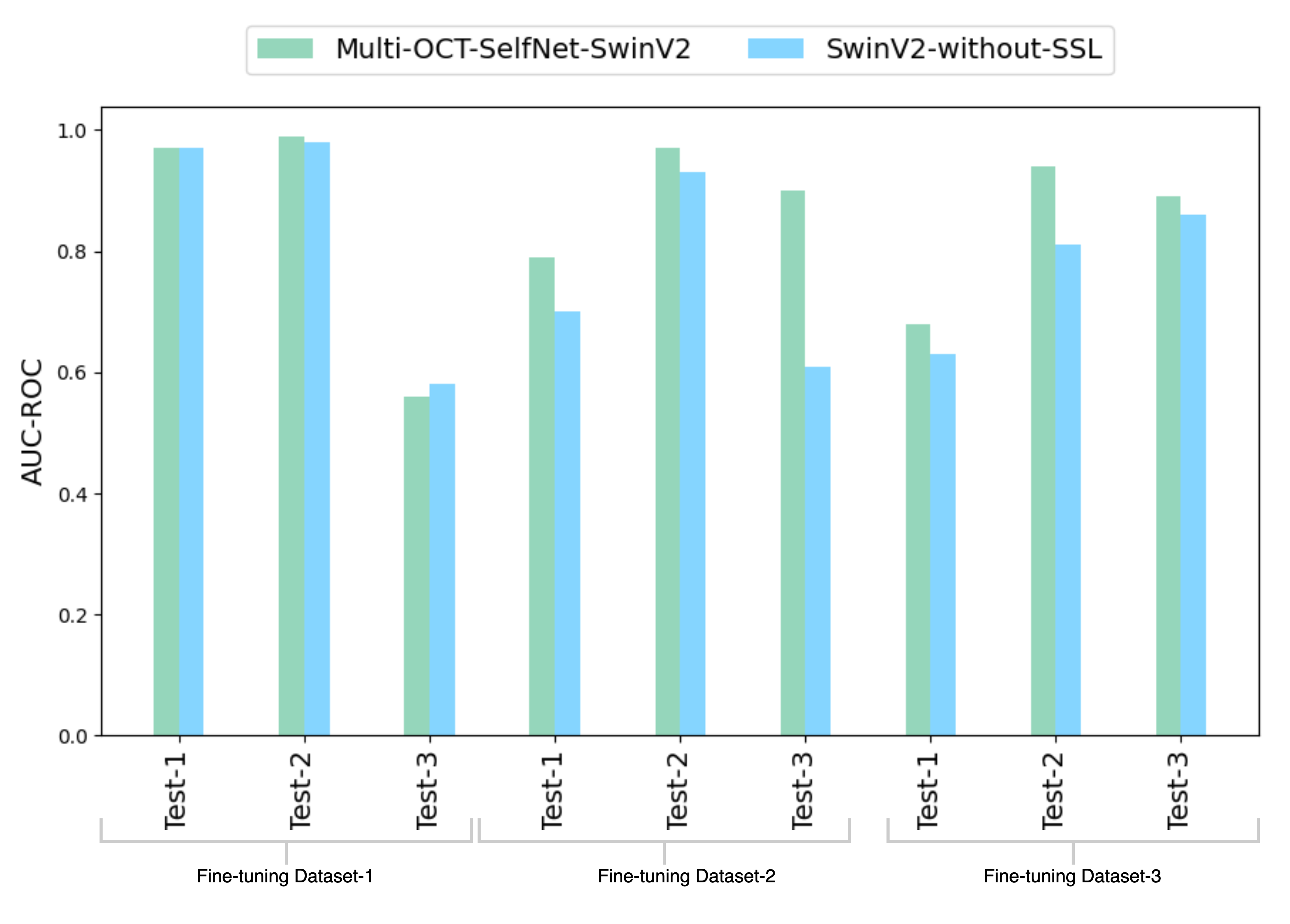}}
\caption{Comparison of AUC-ROC Scores for Multi-OCT-SelfNet-SwinV2 Classifier Without and With Self-Supervised Pretraining Phase Across Three Datasets and Test Sets.
}
\label{fig-table-bar-4}
\end{figure*}

Nevertheless, when evaluated on other test sets for off-domain evaluation, the model exhibited poorer performance in terms of generalization, underscoring the crucial role of self-supervised pre-training in enhancing model robustness and adaptability across diverse datasets. 
The results presented in Table \ref{tab:result-without-ssl} indicate that while the on-domain performance for Dataset-1 remains largely consistent, a slight deterioration is observed in the off-domain evaluation. Conversely, for Dataset-2 and Dataset-3, which are smaller datasets, both the on-domain and off-domain evaluation performances exhibit significant deterioration without the self-supervised step. 
The grouped bar chart in Figure \ref{fig-table-bar-4} compares AUC-ROC scores between self-supervised and non-self-supervised methodologies. Across all three datasets and their corresponding test sets, the non-self-supervised approach consistently shows a decline in performance, highlighting the significant impact of the self-supervised methodology.

To further assess the overall generalization capability, we employed the penalty index, as illustrated in Table \ref{tab:p-score-without-ssl}. This index offers a quantitative measure of the impact of omitting self-supervised pre-training on generalization performance. Notably, the penalty index values are consistently high across most cases when the model is not equipped with pre-trained weights from the self-supervised stage which indicates lower generalization capability. These findings underscore the critical role of self-supervised pre-training in mitigating generalization degradation, particularly evident in scenarios with smaller datasets.

% Please add the following required packages to your document preamble:
% \usepackage{multirow}
% \usepackage[table,xcdraw]{xcolor}
% Beamer presentation requires \usepackage{colortbl} instead of \usepackage[table,xcdraw]{xcolor}
\begin{table}[h]
\centering
\scriptsize
\caption{Analyzing the Impact of Self-Supervised Pre-Training: Comparing Our Framework with SwinV2 Classifier on Test Sets from Three Datasets. Evaluation Includes AUC-ROC, Accuracy,  AUC-PR, and F1-Score for Multi-Class Classification Performance.}
\label{tab:result-without-ssl}
\begin{tabular}{|l|l|lll|lll|lll|lll|}
\hline
 &
  \multicolumn{1}{c|}{} &
  \multicolumn{3}{c|}{AUC-ROC} &
  \multicolumn{3}{c|}{Accuracy} &
  \multicolumn{3}{c|}{AUC-PR} &
  \multicolumn{3}{c|}{F1-Score} \\ \cline{3-14} 
\multirow{-2}{*}{Dataset} &
  \multicolumn{1}{c|}{\multirow{-2}{*}{Classifier Name}} &
  \multicolumn{1}{l|}{Test1} &
  \multicolumn{1}{l|}{Test2} &
  Test3 &
  \multicolumn{1}{l|}{Test1} &
  \multicolumn{1}{l|}{Test2} &
  Test3 &
  \multicolumn{1}{l|}{Test1} &
  \multicolumn{1}{l|}{Test2} &
  Test3 &
  \multicolumn{1}{l|}{Test1} &
  \multicolumn{1}{l|}{Test2} &
  Test3 \\ \hline
\rowcolor[HTML]{F3F3F3} 
\cellcolor[HTML]{F3F3F3} &
  Multi-OCT-SelfNet-SwinV2 &
  
  \multicolumn{1}{l|}{\cellcolor[HTML]{F3F3F3}\textbf{0.97}} &
  \multicolumn{1}{l|}{\cellcolor[HTML]{F3F3F3}\textbf{0.99}} &
  0.56 &
  \multicolumn{1}{l|}{\cellcolor[HTML]{F3F3F3}\textbf{0.90}} &
  \multicolumn{1}{l|}{\cellcolor[HTML]{F3F3F3}\textbf{0.86}} &
   \textbf{0.46} &
  \multicolumn{1}{l|}{\cellcolor[HTML]{F3F3F3}\textbf{0.89}} &
  \multicolumn{1}{l|}{\cellcolor[HTML]{F3F3F3}\textbf{0.98}} &
  \textbf{0.42} &
  \multicolumn{1}{l|}{\cellcolor[HTML]{F3F3F3}\textbf{0.90}} &
  \multicolumn{1}{l|}{\cellcolor[HTML]{F3F3F3}\textbf{0.91}} &
  \textbf{0.87} \\ \cline{2-14} 
\rowcolor[HTML]{F3F3F3} 
\multirow{-2}{*}{\cellcolor[HTML]{F3F3F3}Dataset-1} &
  SwinV2-without-SSL &
  
  \multicolumn{1}{l|}{\cellcolor[HTML]{F3F3F3}\textbf{0.97}} &
  \multicolumn{1}{l|}{\cellcolor[HTML]{F3F3F3}0.98} &
  \textbf{0.58} &
  \multicolumn{1}{l|}{\cellcolor[HTML]{F3F3F3}\textbf{0.90}} &
  \multicolumn{1}{l|}{\cellcolor[HTML]{F3F3F3}0.84} &
  0.40 &
  \multicolumn{1}{l|}{\cellcolor[HTML]{F3F3F3}0.87} &
  \multicolumn{1}{l|}{\cellcolor[HTML]{F3F3F3}0.97} &
  \textbf{0.42} &
  \multicolumn{1}{l|}{\cellcolor[HTML]{F3F3F3}0.89} &
  \multicolumn{1}{l|}{\cellcolor[HTML]{F3F3F3}0.89} &
  0.80 \\ \hline
 &
  \cellcolor[HTML]{F3F3F3}Multi-OCT-SelfNet-SwinV2 &
  
  \multicolumn{1}{l|}{\textbf{0.79}} &
  \multicolumn{1}{l|}{\textbf{0.97}} &
  \textbf{0.90} &
  \multicolumn{1}{l|}{\textbf{0.65}} &
  \multicolumn{1}{l|}{\textbf{0.86}} &
  \textbf{0.45} &
  \multicolumn{1}{l|}{\textbf{0.58}} &
  \multicolumn{1}{l|}{\textbf{0.94}} &
  \textbf{0.70} &
  \multicolumn{1}{l|}{\textbf{0.68}} &
  \multicolumn{1}{l|}{\textbf{0.86}} &
  \textbf{0.54} \\ \cline{2-14} 
\multirow{-2}{*}{Dataset-2} &
  \cellcolor[HTML]{F3F3F3}SwinV2-without-SSL &
  
  \multicolumn{1}{l|}{0.70} &
  \multicolumn{1}{l|}{0.93} &
  0.61 &
  \multicolumn{1}{l|}{0.57} &
  \multicolumn{1}{l|}{0.79} &
  0.39 &
  \multicolumn{1}{l|}{0.47} &
  \multicolumn{1}{l|}{0.85} &
  0.37 &
  \multicolumn{1}{l|}{0.61} &
  \multicolumn{1}{l|}{0.79} &
  0.52 \\ \hline
\rowcolor[HTML]{F3F3F3} 
\cellcolor[HTML]{F3F3F3} &
  Multi-OCT-SelfNet-SwinV2 &
  
  \multicolumn{1}{l|}{\cellcolor[HTML]{F3F3F3}\textbf{0.68}} &
  \multicolumn{1}{l|}{\cellcolor[HTML]{F3F3F3}\textbf{0.94}} &
   \textbf{0.89} &
   \multicolumn{1}{l|}{\cellcolor[HTML]{F3F3F3}0.45} &
  \multicolumn{1}{l|}{\cellcolor[HTML]{F3F3F3}\textbf{0.84}} &
  \textbf{0.86} &
  \multicolumn{1}{l|}{\cellcolor[HTML]{F3F3F3}\textbf{0.49}} &
  \multicolumn{1}{l|}{\cellcolor[HTML]{F3F3F3}\textbf{0.93}} &
  \textbf{0.75} &
  \multicolumn{1}{l|}{\cellcolor[HTML]{F3F3F3}\textbf{0.49}} &
  \multicolumn{1}{l|}{\cellcolor[HTML]{F3F3F3}\textbf{0.84}} &
  \textbf{0.83} \\ \cline{2-14} 
\rowcolor[HTML]{F3F3F3} 
\multirow{-2}{*}{\cellcolor[HTML]{F3F3F3}Dataset-3} &
  SwinV2-without-SSL &
  
  \multicolumn{1}{l|}{\cellcolor[HTML]{F3F3F3}0.63} &
  \multicolumn{1}{l|}{\cellcolor[HTML]{F3F3F3}0.81} &
  0.86 &
  \multicolumn{1}{l|}{\cellcolor[HTML]{F3F3F3}\textbf{0.46}} &
  \multicolumn{1}{l|}{\cellcolor[HTML]{F3F3F3}0.63} &
  0.80 &
  \multicolumn{1}{l|}{\cellcolor[HTML]{F3F3F3}0.45} &
  \multicolumn{1}{l|}{\cellcolor[HTML]{F3F3F3}0.66} &
  0.70 &
  \multicolumn{1}{l|}{\cellcolor[HTML]{F3F3F3}0.45} &
  \multicolumn{1}{l|}{\cellcolor[HTML]{F3F3F3}0.58} &
  0.76 \\ \hline
\end{tabular}
\end{table}

% Please add the following required packages to your document preamble:
% \usepackage{multirow}
% \usepackage[table,xcdraw]{xcolor}
% Beamer presentation requires \usepackage{colortbl} instead of \usepackage[table,xcdraw]{xcolor}
\begin{table}[h]
\centering
\scriptsize
\caption{Analyzing Model's Generalization Performance with or without SSL Pre-training: Comparing Our Proposed Framework with SwinV2 Network which is not pre-trained with SSL, Assessing Penalty-Based Performance Scores Across Three Test Sets for Different Datasets.}
\label{tab:p-score-without-ssl}
\begin{tabular}{|l|l|l|l|l|l|}
\hline
 &
  \multicolumn{1}{c|}{} &
  \multicolumn{1}{c|}{} &
  \multicolumn{1}{c|}{} &
  \multicolumn{1}{c|}{} &
  \multicolumn{1}{c|}{} \\
\multirow{-2}{*}{Dataset} &
  \multicolumn{1}{c|}{\multirow{-2}{*}{Classifier Name}} &
  \multicolumn{1}{c|}{\multirow{-2}{*}{P-Index of AUC-ROC}} &
  \multicolumn{1}{c|}{\multirow{-2}{*}{P-Index of Accuracy}} &
  \multicolumn{1}{c|}{\multirow{-2}{*}{P-Index of AUC-PR}} &
  \multicolumn{1}{c|}{\multirow{-2}{*}{P-Index of F1-Score}} \\ \hline
\rowcolor[HTML]{F3F3F3} 
\cellcolor[HTML]{F3F3F3} &
  Multi-OCT-SelfNet-SwinV2 &
  
  16.0 &
  \textbf{26.0} &
  \textbf{23.67} &
  \textbf{10.67}  \\ \cline{2-6} 
\rowcolor[HTML]{F3F3F3} 
\multirow{-2}{*}{\cellcolor[HTML]{F3F3F3}Dataset-1} &
  SwinV2-without-SSL &
  
  \textbf{15.67} &
  28.67 &
  24.67 &
  14.00 \\ \hline
 &
  \cellcolor[HTML]{F3F3F3}Multi-OCT-SelfNet-SwinV2 &
  
  \textbf{11.33} &
  \textbf{34.67} &
  \textbf{26.00} &
  \textbf{30.66} \\ \cline{2-6} 
\multirow{-2}{*}{Dataset-2} &
  \cellcolor[HTML]{F3F3F3}SwinV2-without-SSL &
  
  25.33 &
  41.67 &
  43.67 &
  36.00 \\ \hline
\rowcolor[HTML]{F3F3F3} 
\cellcolor[HTML]{F3F3F3} &
  Multi-OCT-SelfNet-SwinV2 &
  
  \textbf{16.33} &
  \textbf{28.33} &
  \textbf{27.67} &
  \textbf{28.00} \\ \cline{2-6} 
\rowcolor[HTML]{F3F3F3} 
\multirow{-2}{*}{\cellcolor[HTML]{F3F3F3}Dataset-3} &
  SwinV2-without-SSL &
  
  23.33 &
  36.99 &
  39.67 &
  40.33 \\ \hline
\end{tabular}
\end{table}

\subsubsection{Performance Comparison in Limited Data Settings}

In this experiment, we conducted an in-depth analysis to assess the effectiveness of our proposed framework in scenarios characterized by limited data availability, where acquiring a larger labeled dataset is not feasible. We compared the result to the baseline model. We fine-tuned the model during the fine-tuning stage using only 50\% of the available training data for each dataset, as opposed to utilizing the entire dataset. Subsequently, we evaluated the performance of our proposed framework on the respective test sets.  This experimental setup enabled us to gain valuable insights into the practical applicability of our model in clinical contexts, where data limitations are common and resource constraints pose significant challenges. The analysis from Table \ref{tab:result-half-data} shows notable distinctions across different datasets. Dataset-1 demonstrates better performance with the baseline model, both in on-domain and off-domain evaluations. This outcome is attributed to the dataset's ample size, facilitating robust training. Conversely, Dataset-2 and Dataset-3, which have smaller sample sizes, experienced a significant reduction in performance with the baseline model when the training set size was halved. In these two instances, our proposed method exhibited substantial improvement over the baseline model in both on-domain and off-domain evaluations.

\begin{figure}
\centerline{\includegraphics[width=.75\linewidth]{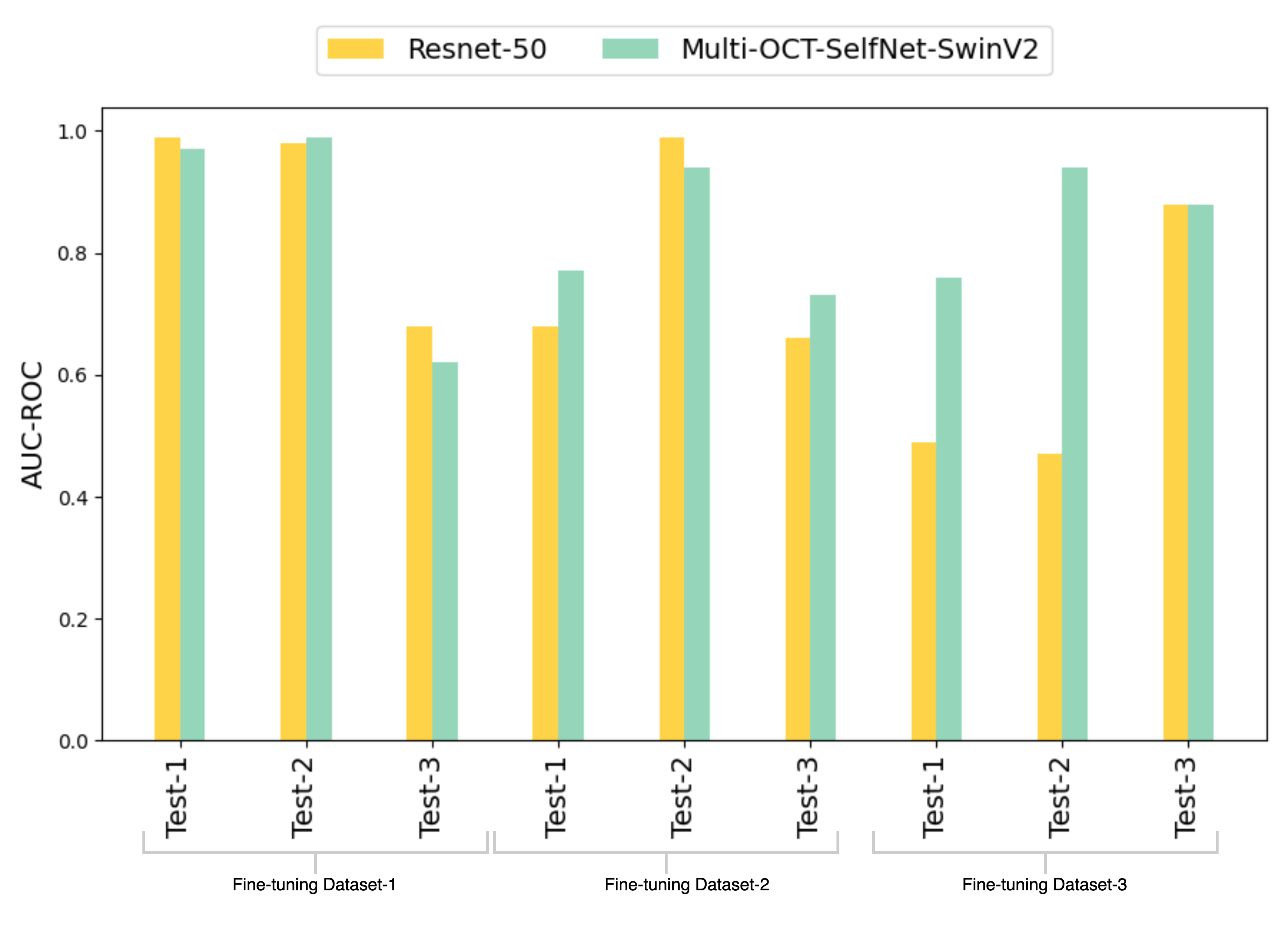}}
\caption{Comparison of AUC-ROC Scores Between Our Method (Multi-OCT-SelfNet-SwinV2) and Baseline (ResNet-50) on Test Sets from Three Datasets, when only 50\% of Training Data has been used for Fine-Tuning. }
\label{fig-table-bar-6}
\end{figure}

The grouped bar chart in Figure \ref{fig-table-bar-6} compares AUC-ROC scores between the baseline model and our proposed model in this limited data-settings experiment. The chart further validates the superior performance of our proposed Multi-OCT-SelfNet-SwinV2 model, as it consistently shows higher bar heights compared to the baseline model in most cases.

To further quantify the model's generalization capabilities, Table \ref{tab:p-index-half-data} provides insights into the penalty scores. Remarkably, our proposed model demonstrated significantly lower penalty scores compared to the baseline model for Dataset-2 and Dataset-3, underscoring its superior ability to generalize across datasets even with limited samples. 

These findings highlight the efficacy of our proposed method, particularly in scenarios with smaller datasets, where it outperforms the baseline model and showcases enhanced generalization capabilities.

% Please add the following required packages to your document preamble:
% \usepackage{multirow}
% \usepackage[table,xcdraw]{xcolor}
% Beamer presentation requires \usepackage{colortbl} instead of \usepackage[table,xcdraw]{xcolor}
\begin{table}[h]
\centering
\scriptsize
\caption{Analyzing the Performance in Limited Data Settings: Comparison of our work with the baseline methods on test sets from three datasets, using only 50\% of the training set from each dataset in finetuning. The evaluation focuses on the classification AUC-ROC, accuracy,  AUC-PR, and F1-score.}
\label{tab:result-half-data}
\begin{tabular}{|l|l|lll|lll|lll|lll|}
\hline
 &
  \multicolumn{1}{c|}{} &
  \multicolumn{3}{c|}{AUC-ROC} &
  \multicolumn{3}{c|}{Accuracy} &
  \multicolumn{3}{c|}{AUC-PR} &
  \multicolumn{3}{c|}{F1-Score} \\ \cline{3-14} 
\multirow{-2}{*}{Dataset} &
  \multicolumn{1}{c|}{\multirow{-2}{*}{Classifier Name}} &
  \multicolumn{1}{l|}{Test1} &
  \multicolumn{1}{l|}{Test2} &
  Test3 &
  \multicolumn{1}{l|}{Test1} &
  \multicolumn{1}{l|}{Test2} &
  Test3 &
  \multicolumn{1}{l|}{Test1} &
  \multicolumn{1}{l|}{Test2} &
  Test3 &
  \multicolumn{1}{l|}{Test1} &
  \multicolumn{1}{l|}{Test2} &
  Test3 \\ \hline
\cellcolor[HTML]{F3F3F3} &
  \cellcolor[HTML]{F3F3F3}Resnet-50 &
  
  \multicolumn{1}{l|}{0.99} &
  \multicolumn{1}{l|}{0.98} &
  0.68 &
  \multicolumn{1}{l|}{0.94} &
  \multicolumn{1}{l|}{0.91} &
  0.66 &
  \multicolumn{1}{l|}{0.96} &
  \multicolumn{1}{l|}{0.98} &
  0.44 &
  \multicolumn{1}{l|}{0.94} &
  \multicolumn{1}{l|}{0.91} &
  0.61 \\ \cline{2-14} 
\multirow{-2}{*}{\cellcolor[HTML]{F3F3F3}Dataset-1} &
  Multi-OCT-SelfNet-SwinV2 &
  
  \multicolumn{1}{l|}{0.97} &
  \multicolumn{1}{l|}{0.99} &
   0.62&
   \multicolumn{1}{l|}{0.90} &
  \multicolumn{1}{l|}{0.89} &
   0.46&
  \multicolumn{1}{l|}{0.89} &
  \multicolumn{1}{l|}{0.98} &
   0.44&
  \multicolumn{1}{l|}{0.90} &
  \multicolumn{1}{l|}{0.93} &
   0.61\\ \hline
\cellcolor[HTML]{F3F3F3} &
  \cellcolor[HTML]{F3F3F3}Resnet-50 &
  
  \multicolumn{1}{l|}{0.68} &
  \multicolumn{1}{l|}{0.99} &
  0.66 &
  \multicolumn{1}{l|}{0.30} &
  \multicolumn{1}{l|}{0.90} &
  0.0 &
  \multicolumn{1}{l|}{0.40} &
  \multicolumn{1}{l|}{0.99} &
  0.59 &
  \multicolumn{1}{l|}{0.33} &
  \multicolumn{1}{l|}{0.90} &
  0.0 \\ \cline{2-14} 
\multirow{-2}{*}{\cellcolor[HTML]{F3F3F3}Dataset-2} &
  Multi-OCT-SelfNet-SwinV2 &
  
  \multicolumn{1}{l|}{0.77} &
  \multicolumn{1}{l|}{0.94} &
  0.73 &
  \multicolumn{1}{l|}{0.70} &
  \multicolumn{1}{l|}{0.83} &
  0.41 &
  \multicolumn{1}{l|}{0.56} &
  \multicolumn{1}{l|}{0.90} &
  0.48 &
  \multicolumn{1}{l|}{0.71} &
  \multicolumn{1}{l|}{0.83} &
  0.62 \\ \hline
\cellcolor[HTML]{F3F3F3} &
  \cellcolor[HTML]{F3F3F3}Resnet-50 &
  
  \multicolumn{1}{l|}{0.49} &
  \multicolumn{1}{l|}{0.47} &
  0.88 &
  \multicolumn{1}{l|}{0.37} &
  \multicolumn{1}{l|}{0.0} &
  0.91 &
  \multicolumn{1}{l|}{0.43} &
  \multicolumn{1}{l|}{0.56} &
  0.72 &
  \multicolumn{1}{l|}{0.20} &
  \multicolumn{1}{l|}{0.0} &
  0.87 \\ \cline{2-14} 
\multirow{-2}{*}{\cellcolor[HTML]{F3F3F3}Dataset-3} &
  \cellcolor[HTML]{F3F3F3}Multi-OCT-SelfNet-SwinV2 &
  
  \multicolumn{1}{l|}{0.76} &
  \multicolumn{1}{l|}{0.94} &
  0.88 &
  \multicolumn{1}{l|}{0.47} &
  \multicolumn{1}{l|}{0.80} &
  0.81 &
  \multicolumn{1}{l|}{0.56} &
  \multicolumn{1}{l|}{0.91} &
  0.73 &
  \multicolumn{1}{l|}{0.49} &
  \multicolumn{1}{l|}{0.83} &
  0.78 \\ \hline
\end{tabular}
\end{table}

% Please add the following required packages to your document preamble:
% \usepackage{multirow}
% \usepackage[table,xcdraw]{xcolor}
% Beamer presentation requires \usepackage{colortbl} instead of \usepackage[table,xcdraw]{xcolor}
\begin{table}[h]
\centering
\scriptsize
\caption{Analyzing Model’s Generalization Performance in Limited Data Settings: Comparing Our Proposed Framework with SwinV2 Network With the Baseline Model on test
sets from three datasets, using only 50\% of the training set from each dataset in fine-tuning. Assessing Penalty-Based Performance Scores Across Three Test Sets for Different Datasets.}
\label{tab:p-index-half-data}
\begin{tabular}{|l|l|l|l|l|l|}
\hline
 &
  \multicolumn{1}{c|}{} &
  \multicolumn{1}{c|}{} &
  \multicolumn{1}{c|}{} &
  \multicolumn{1}{c|}{} &
  \multicolumn{1}{c|}{} \\
\multirow{-2}{*}{Dataset} &
  \multicolumn{1}{c|}{\multirow{-2}{*}{Classifier Name}} &
  \multicolumn{1}{c|}{\multirow{-2}{*}{P-Index of AUC-ROC}} &
  \multicolumn{1}{c|}{\multirow{-2}{*}{P-Index of Accuracy}} &
  \multicolumn{1}{c|}{\multirow{-2}{*}{P-Index of AUC-PR}} &
  \multicolumn{1}{c|}{\multirow{-2}{*}{P-Index of F1-Score}} \\ \hline
\rowcolor[HTML]{F3F3F3} 
\cellcolor[HTML]{F3F3F3} &
  Resnet-50 &
  
  \textbf{11.67} &
  \textbf{16.33} &
  \textbf{20.67} &
  \textbf{18.00} \\ \cline{2-6} 
\rowcolor[HTML]{F3F3F3} 
\multirow{-2}{*}{\cellcolor[HTML]{F3F3F3}Dataset-1} &
  Multi-OCT-SelfNet-SwinV2 &
  
  14.00 &
  25.0 &
  23.0 &
  18.67 \\ \hline
 &
  \cellcolor[HTML]{F3F3F3}Resnet-50 &
  
  22.33 &
  60.0 &
  \textbf{34.0} &
  59.0 \\ \cline{2-6} 
\multirow{-2}{*}{Dataset-2} &
  \cellcolor[HTML]{F3F3F3}Multi-OCT-SelfNet-SwinV2 &
  
  \textbf{18.67} &
  \textbf{35.33} &
  35.33 &
  \textbf{28.00} \\ \hline
\rowcolor[HTML]{F3F3F3} 
\cellcolor[HTML]{F3F3F3} &
  Resnet-50 &
  
  38.67 &
  57.33 &
  43.0 &
  64.33 \\ \cline{2-6} 
\rowcolor[HTML]{F3F3F3} 
\multirow{-2}{*}{\cellcolor[HTML]{F3F3F3}Dataset-3} &
  Multi-OCT-SelfNet-SwinV2 &
  
  \textbf{14.00} &
  \textbf{30.67} &
  \textbf{26.67} &
  \textbf{30.0} \\ \hline
\end{tabular}
\end{table}

\begin{figure}
\centerline{\includegraphics[width=.75\linewidth]{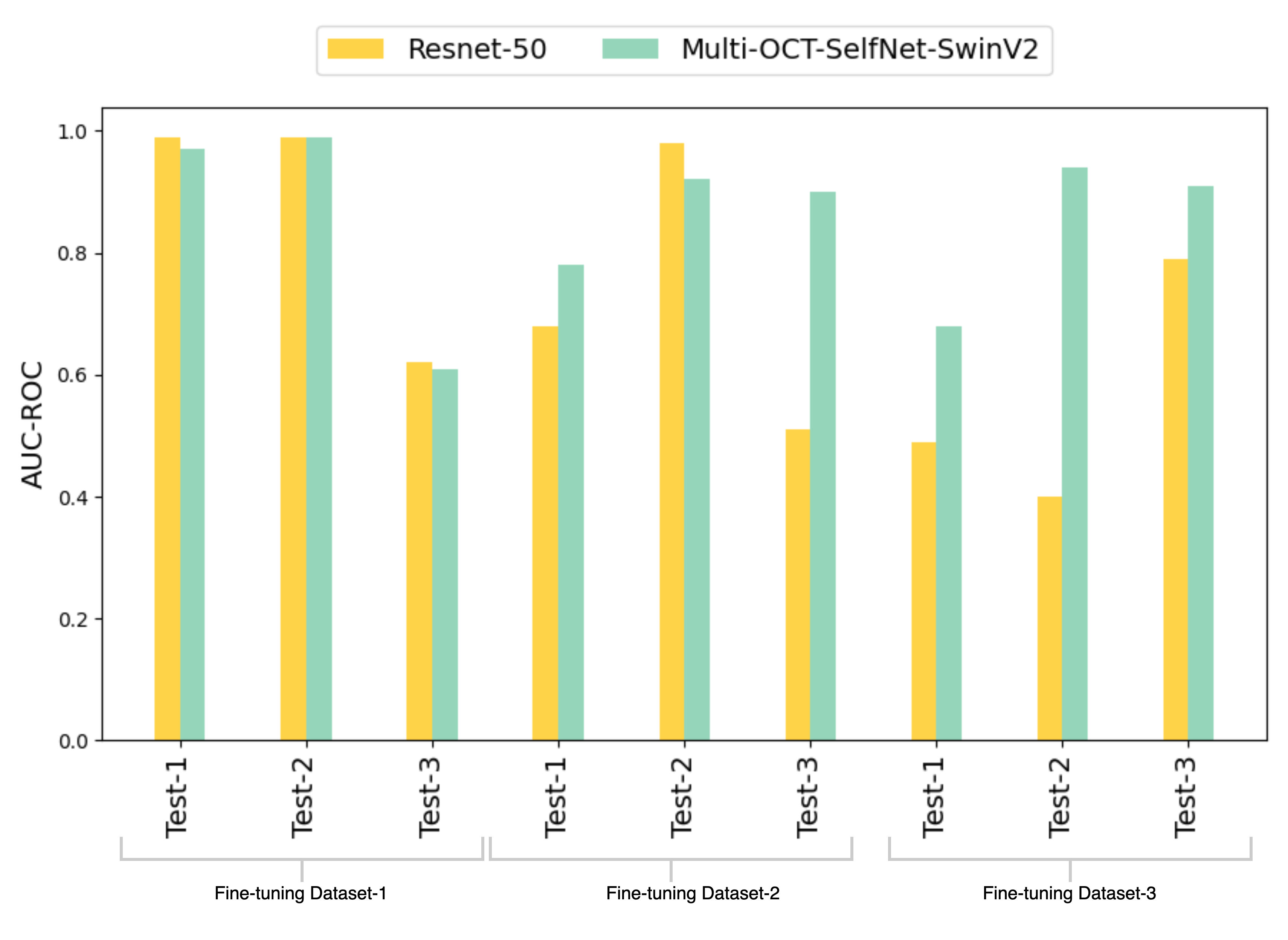}}
\caption{Comparison of AUC-ROC Scores Between Our Method (Multi-OCT-SelfNet-SwinV2) and Baseline (ResNet-50) on Test Sets from Three Datasets with class imbalance correction.}
\label{fig-table-bar-8}
\end{figure}

\red{ \subsubsection{Performance Evaluation on the Effect of Correcting Data Imbalance}
In this experiment, we have addressed the data imbalance issue by assigning different weights to each class in the loss function. This strategy ensures that the model focuses more on the minority class during training, reducing the bias towards the majority class. We compared the results of our model, using this weight-adjusting method, to the baseline model. The goal was to observe how our model performs compared to the baseline model when correcting data imbalance issues. Table \ref{tab:result-correcting}  provides detailed results with accuracy, AUC-ROC, AUC-PR, and F1-score. The bar chart in Figure \ref{fig-table-bar-8} illustrates a similar trend to the previous experiments, when the dataset is sufficiently large (as with DS-1), the performance of both models is comparable. However, as the dataset size decreases, a more significant performance difference emerges, with our proposed model outperforming the baseline in most cases, particularly in the smallest dataset, DS-3. Additionally, Table \ref{tab:table-bar-p-index-correct} shows that the P-Index is lower for our proposed model in most cases, indicating its superior generalization capability compared to the baseline. }

\red{When comparing these results to the previous experiment, where class imbalance was not addressed, as shown in Table \ref{tab:result-encoder}, it is evident that correcting the imbalance causes a slight decline in performance for both models. Despite this reduction, our proposed model still outperforms the baseline in most cases, achieving a higher score between the two. This indicates that even with the trade-off of a minor performance drop, our model has a notable advantage in overall performance.}

% Please add the following required packages to your document preamble:
% \usepackage{multirow}
% \usepackage[table,xcdraw]{xcolor}
% Beamer presentation requires \usepackage{colortbl} instead of \usepackage[table,xcdraw]{xcolor}
\begin{table}[]
\centering
\scriptsize
\caption{Analyzing the Performance Class Imbalance Correction: Comparison of our framework with baseline methods (ResNet-50) on the test sets from three datasets.}
\label{tab:result-correcting}
\begin{tabular}{|l|l|lll|lll|lll|lll|}
\hline
\rowcolor[HTML]{FFFFFF} 
\cellcolor[HTML]{FFFFFF} &
  \cellcolor[HTML]{FFFFFF} &
  \multicolumn{3}{c|}{\cellcolor[HTML]{FFFFFF}AUC-ROC} &
  \multicolumn{3}{c|}{\cellcolor[HTML]{FFFFFF}Accuracy} &
  \multicolumn{3}{c|}{\cellcolor[HTML]{FFFFFF}AUC-PR} &
  \multicolumn{3}{c|}{\cellcolor[HTML]{FFFFFF}F1-Score} \\ \cline{3-14} 
\rowcolor[HTML]{FFFFFF} 
\multirow{-2}{*}{\cellcolor[HTML]{FFFFFF}Train Set} &
  \multirow{-2}{*}{\cellcolor[HTML]{FFFFFF}Classifier Name} &
  \multicolumn{1}{c|}{\cellcolor[HTML]{FFFFFF}Test-1} &
  \multicolumn{1}{c|}{\cellcolor[HTML]{FFFFFF}Test-2} &
  \multicolumn{1}{c|}{\cellcolor[HTML]{FFFFFF}Test-3} &
  \multicolumn{1}{c|}{\cellcolor[HTML]{FFFFFF}Test-1} &
  \multicolumn{1}{c|}{\cellcolor[HTML]{FFFFFF}Test-2} &
  \multicolumn{1}{c|}{\cellcolor[HTML]{FFFFFF}Test-3} &
  \multicolumn{1}{c|}{\cellcolor[HTML]{FFFFFF}Test-1} &
  \multicolumn{1}{c|}{\cellcolor[HTML]{FFFFFF}Test-2} &
  \multicolumn{1}{c|}{\cellcolor[HTML]{FFFFFF}Test-3} &
  \multicolumn{1}{c|}{\cellcolor[HTML]{FFFFFF}Test-1} &
  \multicolumn{1}{c|}{\cellcolor[HTML]{FFFFFF}Test-2} &
  \multicolumn{1}{c|}{\cellcolor[HTML]{FFFFFF}Test-3} \\ \hline
\rowcolor[HTML]{F3F3F3} 
\cellcolor[HTML]{FFFFFF} &
  \cellcolor[HTML]{FFFFFF}Resnet50 &
  \multicolumn{1}{l|}{\cellcolor[HTML]{F3F3F3}0.99} &
  \multicolumn{1}{l|}{\cellcolor[HTML]{F3F3F3}0.99} &
  0.62 &
  \multicolumn{1}{l|}{\cellcolor[HTML]{F3F3F3}0.94} &
  \multicolumn{1}{l|}{\cellcolor[HTML]{F3F3F3}0.93} &
  0.14 &
  \multicolumn{1}{l|}{\cellcolor[HTML]{F3F3F3}0.96} &
  \multicolumn{1}{l|}{\cellcolor[HTML]{F3F3F3}0.98} &
  0.41 &
  \multicolumn{1}{l|}{\cellcolor[HTML]{F3F3F3}0.94} &
  \multicolumn{1}{l|}{\cellcolor[HTML]{F3F3F3}0.93} &
  0.09 \\ \cline{2-14} 
\rowcolor[HTML]{F3F3F3} 
\multirow{-2}{*}{\cellcolor[HTML]{FFFFFF}DS-1} &
  \cellcolor[HTML]{FFFFFF}Multi-OCT-SelfNet-SwinV2 &
  \multicolumn{1}{l|}{\cellcolor[HTML]{F3F3F3}0.97} &
  \multicolumn{1}{l|}{\cellcolor[HTML]{F3F3F3}0.99} &
  0.61 &
  \multicolumn{1}{l|}{\cellcolor[HTML]{F3F3F3}0.86} &
  \multicolumn{1}{l|}{\cellcolor[HTML]{F3F3F3}0.92} &
  0.31 &
  \multicolumn{1}{l|}{\cellcolor[HTML]{F3F3F3}0.86} &
  \multicolumn{1}{l|}{\cellcolor[HTML]{F3F3F3}0.99} &
  0.44 &
  \multicolumn{1}{l|}{\cellcolor[HTML]{F3F3F3}0.87} &
  \multicolumn{1}{l|}{\cellcolor[HTML]{F3F3F3}0.93} &
  0.61 \\ \hline
\rowcolor[HTML]{FFFFFF} 
\cellcolor[HTML]{FFFFFF} &
  Resnet50 &
  \multicolumn{1}{l|}{\cellcolor[HTML]{FFFFFF}0.68} &
  \multicolumn{1}{l|}{\cellcolor[HTML]{FFFFFF}0.98} &
  0.51 &
  \multicolumn{1}{l|}{\cellcolor[HTML]{FFFFFF}0.35} &
  \multicolumn{1}{l|}{\cellcolor[HTML]{FFFFFF}0.86} &
  0.04 &
  \multicolumn{1}{l|}{\cellcolor[HTML]{FFFFFF}0.40} &
  \multicolumn{1}{l|}{\cellcolor[HTML]{FFFFFF}0.95} &
  0.52 &
  \multicolumn{1}{l|}{\cellcolor[HTML]{FFFFFF}0.40} &
  \multicolumn{1}{l|}{\cellcolor[HTML]{FFFFFF}0.86} &
  0.02 \\ \cline{2-14} 
\rowcolor[HTML]{FFFFFF} 
\multirow{-2}{*}{\cellcolor[HTML]{FFFFFF}DS-2} &
  Multi-OCT-SelfNet-SwinV2 &
  \multicolumn{1}{l|}{\cellcolor[HTML]{FFFFFF}0.78} &
  \multicolumn{1}{l|}{\cellcolor[HTML]{FFFFFF}0.92} &
  0.90 &
  \multicolumn{1}{l|}{\cellcolor[HTML]{FFFFFF}0.65} &
  \multicolumn{1}{l|}{\cellcolor[HTML]{FFFFFF}0.78} &
  0.37 &
  \multicolumn{1}{l|}{\cellcolor[HTML]{FFFFFF}0.56} &
  \multicolumn{1}{l|}{\cellcolor[HTML]{FFFFFF}0.84} &
  0.69 &
  \multicolumn{1}{l|}{\cellcolor[HTML]{FFFFFF}0.67} &
  \multicolumn{1}{l|}{\cellcolor[HTML]{FFFFFF}0.78} &
  0.80 \\ \hline
\rowcolor[HTML]{F3F3F3} 
\cellcolor[HTML]{FFFFFF} &
  \cellcolor[HTML]{FFFFFF}Resnet50 &
  \multicolumn{1}{l|}{\cellcolor[HTML]{F3F3F3}0.49} &
  \multicolumn{1}{l|}{\cellcolor[HTML]{F3F3F3}0.40} &
  0.79 &
  \multicolumn{1}{l|}{\cellcolor[HTML]{F3F3F3}0.37} &
  \multicolumn{1}{l|}{\cellcolor[HTML]{F3F3F3}0.0} &
  0.66 &
  \multicolumn{1}{l|}{\cellcolor[HTML]{F3F3F3}0.42} &
  \multicolumn{1}{l|}{\cellcolor[HTML]{F3F3F3}0.63} &
  0.57 &
  \multicolumn{1}{l|}{\cellcolor[HTML]{F3F3F3}0.20} &
  \multicolumn{1}{l|}{\cellcolor[HTML]{F3F3F3}0.0} &
  0.65 \\ \cline{2-14} 
\rowcolor[HTML]{F3F3F3} 
\multirow{-2}{*}{\cellcolor[HTML]{FFFFFF}DS-3} &
  \cellcolor[HTML]{FFFFFF}Multi-OCT-SelfNet-SwinV2 &
  \multicolumn{1}{l|}{\cellcolor[HTML]{F3F3F3}0.68} &
  \multicolumn{1}{l|}{\cellcolor[HTML]{F3F3F3}0.94} &
  0.91 &
  \multicolumn{1}{l|}{\cellcolor[HTML]{F3F3F3}0.32} &
  \multicolumn{1}{l|}{\cellcolor[HTML]{F3F3F3}0.62} &
  0.76 &
  \multicolumn{1}{l|}{\cellcolor[HTML]{F3F3F3}0.49} &
  \multicolumn{1}{l|}{\cellcolor[HTML]{F3F3F3}0.91} &
  0.79 &
  \multicolumn{1}{l|}{\cellcolor[HTML]{F3F3F3}0.55} &
  \multicolumn{1}{l|}{\cellcolor[HTML]{F3F3F3}0.80} &
  0.76 \\ \hline
\end{tabular}
\end{table}

% Please add the following required packages to your document preamble:
% \usepackage{multirow}
% \usepackage[table,xcdraw]{xcolor}
% Beamer presentation requires \usepackage{colortbl} instead of \usepackage[table,xcdraw]{xcolor}
\begin{table}[]
\centering
\scriptsize
\caption{Analyzing Model’s Generalization Performance with Class Imbalance Correction: Comparison of our framework with baseline methods (ResNet-50) on the test sets from three datasets.}
\label{tab:table-bar-p-index-correct}
\begin{tabular}{|l|l|l|l|l|l|}
\hline
 &
  \multicolumn{1}{c|}{} &
  \multicolumn{1}{c|}{} &
  \multicolumn{1}{c|}{} &
  \multicolumn{1}{c|}{} &
  \multicolumn{1}{c|}{} \\
\multirow{-2}{*}{Dataset} &
  \multicolumn{1}{c|}{\multirow{-2}{*}{Classifier Name}} &
  \multicolumn{1}{c|}{\multirow{-2}{*}{P-Index of Accuracy}} &
  \multicolumn{1}{c|}{\multirow{-2}{*}{P-Index of AUC-ROC}} &
  \multicolumn{1}{c|}{\multirow{-2}{*}{P-Index of AUC-PR}} &
  \multicolumn{1}{c|}{\multirow{-2}{*}{P-Index of F1-Score}} \\ \hline
 &
  \cellcolor[HTML]{F3F3F3}Resnet-50 &
  \cellcolor[HTML]{F3F3F3}33.0 &
  \cellcolor[HTML]{F3F3F3}\textbf{13.33} &
  \cellcolor[HTML]{F3F3F3}\textbf{21.67} &
  \cellcolor[HTML]{F3F3F3}34.67 \\ \cline{2-6} 
\multirow{-2}{*}{Dataset-1} &
  \cellcolor[HTML]{F3F3F3}Multi-OCT-SelfNet-SwinV2 &
  \cellcolor[HTML]{F3F3F3}\textbf{30.33} &
  \cellcolor[HTML]{F3F3F3}14.33 &
  \cellcolor[HTML]{F3F3F3}23.67 &
  \cellcolor[HTML]{F3F3F3}\textbf{19.67} \\ \hline
 &
  \cellcolor[HTML]{F3F3F3}Resnet-50 &
  27.67 &
  58.33 &
  37.67 &
  57.33 \\ \cline{2-6} 
\multirow{-2}{*}{Dataset-2} &
  \cellcolor[HTML]{F3F3F3}Multi-OCT-SelfNet-SwinV2 &
  \textbf{13.33} &
  \textbf{40.0} &
  \textbf{30.33} &
  \textbf{25.0} \\ \hline
 &
  \cellcolor[HTML]{F3F3F3}Resnet-50 &
  \cellcolor[HTML]{F3F3F3}44.0 &
  \cellcolor[HTML]{F3F3F3}65.67 &
  \cellcolor[HTML]{F3F3F3}46.0 &
  \cellcolor[HTML]{F3F3F3}71.67 \\ \cline{2-6} 
\multirow{-2}{*}{Dataset-3} &
  \cellcolor[HTML]{F3F3F3}Multi-OCT-SelfNet-SwinV2 &
  \cellcolor[HTML]{F3F3F3}\textbf{15.67} &
  \cellcolor[HTML]{F3F3F3}\textbf{43.33} &
  \cellcolor[HTML]{F3F3F3}\textbf{27.0} &
  \cellcolor[HTML]{F3F3F3}\textbf{29.67} \\ \hline
\end{tabular}
\end{table}
\section{Future Work}
\red {While our framework has shown significant potential in the automated classification of retinal diseases using OCT images, its complexity poses challenges in terms of interpretability. As we move forward, enhancing the transparency of our model is a key priority. By improving interpretability, we aim to build AI-based diagnostic tools that clinicians can trust and readily understand, ensuring that these models are not only powerful but also transparent in their operations. Additionally, we recognize the importance of continuous improvement in AI systems, particularly in the context of their application in diverse clinical environments. Our future work will focus on integrating a human-in-the-loop system to complement the model’s capabilities. By involving human expertise, we can enhance the model’s adaptability, enabling it to learn and improve from real-world feedback continuously. This approach will help mitigate issues related to cross-domain generalization, ensuring that our model maintains high performance across different clinical settings. }

\red{Inconsistent labeling or errors in the dataset can introduce biases and negatively affect the model’s performance, particularly when combining datasets from multiple sources. In this process, the domain experts will periodically review the model’s predictions, especially in examples where the model is uncertain. This iterative feedback loop will help to continuously refine label quality and enhance model accuracy by effectively managing labeling inconsistencies and errors in the dataset.}

\section{Conclusion}
The proposed work aims to leverage innovative techniques such as multi-source data fusion, self-supervised learning, and transformer networks to overcome the challenges posed by limited data availability in retinal disease diagnosis. During the self-supervised phase, the model is pre-trained using publicly available datasets, ensuring that sensitive clinical data is not exposed. In the downstream classification task specific to individual clinical settings, the model is fine-tuned solely on the local dataset, eliminating the need to share any private data. This approach preserves data privacy while allowing the model to adapt effectively to diverse clinical environments. This framework will be beneficial in situations where access to extensive datasets is limited offering a scalable and practical approach to implementing medical AI solutions. The results of our study showcase the effectiveness of our proposed framework, Multi-OCT-SelfNet, which consistently surpasses the baseline performance. Our approach demonstrates superior performance scores across all datasets, particularly excelling with smaller datasets. Through meticulous experimentation, we've validated the efficacy of our methodology. The incorporation of data fusion and self-supervised pre-training significantly enhances performance, as evidenced by our ablation study. This highlights the significance of these components in improving both the resilience and accuracy of our model, thereby facilitating robust generalization.

%%
%% The next two lines define the bibliography style to be used, and
%% the bibliography file.
\bibliographystyle{ACM-Reference-Format}
\bibliography{egbib}

%%% -*-BibTeX-*-
%%% Do NOT edit. File created by BibTeX with style
%%% ACM-Reference-Format-Journals [18-Jan-2012].

\begin{thebibliography}{52}

%%% ====================================================================
%%% NOTE TO THE USER: you can override these defaults by providing
%%% customized versions of any of these macros before the \bibliography
%%% command.  Each of them MUST provide its own final punctuation,
%%% except for \shownote{}, \showDOI{}, and \showURL{}.  The latter two
%%% do not use final punctuation, in order to avoid confusing it with
%%% the Web address.
%%%
%%% To suppress output of a particular field, define its macro to expand
%%% to an empty string, or better, \unskip, like this:
%%%
%%% \newcommand{\showDOI}[1]{\unskip}   % LaTeX syntax
%%%
%%% \def \showDOI #1{\unskip}           % plain TeX syntax
%%%
%%% ====================================================================

\ifx \showCODEN    \undefined \def \showCODEN     #1{\unskip}     \fi
\ifx \showDOI      \undefined \def \showDOI       #1{#1}\fi
\ifx \showISBNx    \undefined \def \showISBNx     #1{\unskip}     \fi
\ifx \showISBNxiii \undefined \def \showISBNxiii  #1{\unskip}     \fi
\ifx \showISSN     \undefined \def \showISSN      #1{\unskip}     \fi
\ifx \showLCCN     \undefined \def \showLCCN      #1{\unskip}     \fi
\ifx \shownote     \undefined \def \shownote      #1{#1}          \fi
\ifx \showarticletitle \undefined \def \showarticletitle #1{#1}   \fi
\ifx \showURL      \undefined \def \showURL       {\relax}        \fi
% The following commands are used for tagged output and should be
% invisible to TeX
\providecommand\bibfield[2]{#2}
\providecommand\bibinfo[2]{#2}
\providecommand\natexlab[1]{#1}
\providecommand\showeprint[2][]{arXiv:#2}

\bibitem[Alam et~al\mbox{.}(2020a)]%
        {Alam:20}
\bibfield{author}{\bibinfo{person}{Minhaj Alam}, \bibinfo{person}{David Le}, \bibinfo{person}{Taeyoon Son}, \bibinfo{person}{Jennifer~I. Lim}, {and} \bibinfo{person}{Xincheng Yao}.} \bibinfo{year}{2020}\natexlab{a}.
\newblock \showarticletitle{AV-Net: deep learning for fully automated artery-vein classification in optical coherence tomography angiography}.
\newblock \bibinfo{journal}{\emph{Biomed. Opt. Express}} \bibinfo{volume}{11}, \bibinfo{number}{9} (\bibinfo{date}{Sep} \bibinfo{year}{2020}), \bibinfo{pages}{5249--5257}.
\newblock
\urldef\tempurl%
\url{https://doi.org/10.1364/BOE.399514}
\showDOI{\tempurl}


\bibitem[Alam et~al\mbox{.}(2020b)]%
        {alam2020quantitative}
\bibfield{author}{\bibinfo{person}{Minhaj Alam}, \bibinfo{person}{Yue Zhang}, \bibinfo{person}{Jennifer~I Lim}, \bibinfo{person}{RVP Chan}, \bibinfo{person}{Min Yang}, {and} \bibinfo{person}{Xincheng Yao}.} \bibinfo{year}{2020}\natexlab{b}.
\newblock \showarticletitle{Quantitative OCT angiography features for objective classification and staging of diabetic retinopathy}.
\newblock \bibinfo{journal}{\emph{Retina (Philadelphia, Pa.)}} (\bibinfo{year}{2020}).
\newblock


\bibitem[Alshammari et~al\mbox{.}(2022)]%
        {alshammari2022olive}
\bibfield{author}{\bibinfo{person}{Hamoud Alshammari}, \bibinfo{person}{Karim Gasmi}, \bibinfo{person}{Ibtihel Ben~Ltaifa}, \bibinfo{person}{Moez Krichen}, \bibinfo{person}{Lassaad Ben~Ammar}, {and} \bibinfo{person}{Mahmood~A Mahmood}.} \bibinfo{year}{2022}\natexlab{}.
\newblock \showarticletitle{Olive disease classification based on vision transformer and CNN models}.
\newblock \bibinfo{journal}{\emph{Computational Intelligence and Neuroscience}}  \bibinfo{volume}{2022} (\bibinfo{year}{2022}).
\newblock


\bibitem[Awais et~al\mbox{.}(2017)]%
        {8120661}
\bibfield{author}{\bibinfo{person}{Muhammad Awais}, \bibinfo{person}{Henning Müller}, \bibinfo{person}{Tong~B. Tang}, {and} \bibinfo{person}{Fabrice Meriaudeau}.} \bibinfo{year}{2017}\natexlab{}.
\newblock \showarticletitle{Classification of SD-OCT images using a Deep learning approach}. In \bibinfo{booktitle}{\emph{2017 IEEE International Conference on Signal and Image Processing Applications (ICSIPA)}}. \bibinfo{pages}{489--492}.
\newblock
\urldef\tempurl%
\url{https://doi.org/10.1109/ICSIPA.2017.8120661}
\showDOI{\tempurl}


\bibitem[Ayana et~al\mbox{.}(2023)]%
        {ayana2023vision}
\bibfield{author}{\bibinfo{person}{Gelan Ayana}, \bibinfo{person}{Kokeb Dese}, \bibinfo{person}{Yisak Dereje}, \bibinfo{person}{Yonas Kebede}, \bibinfo{person}{Hika Barki}, \bibinfo{person}{Dechassa Amdissa}, \bibinfo{person}{Nahimiya Husen}, \bibinfo{person}{Fikadu Mulugeta}, \bibinfo{person}{Bontu Habtamu}, {and} \bibinfo{person}{Se-Woon Choe}.} \bibinfo{year}{2023}\natexlab{}.
\newblock \showarticletitle{Vision-Transformer-Based Transfer Learning for Mammogram Classification}.
\newblock \bibinfo{journal}{\emph{Diagnostics}} \bibinfo{volume}{13}, \bibinfo{number}{2} (\bibinfo{year}{2023}), \bibinfo{pages}{178}.
\newblock


\bibitem[Bressem et~al\mbox{.}(2020)]%
        {bressem2020comparing}
\bibfield{author}{\bibinfo{person}{Keno~K Bressem}, \bibinfo{person}{Lisa~C Adams}, \bibinfo{person}{Christoph Erxleben}, \bibinfo{person}{Bernd Hamm}, \bibinfo{person}{Stefan~M Niehues}, {and} \bibinfo{person}{Janis~L Vahldiek}.} \bibinfo{year}{2020}\natexlab{}.
\newblock \showarticletitle{Comparing different deep learning architectures for classification of chest radiographs}.
\newblock \bibinfo{journal}{\emph{Scientific reports}} \bibinfo{volume}{10}, \bibinfo{number}{1} (\bibinfo{year}{2020}), \bibinfo{pages}{13590}.
\newblock


\bibitem[Carion et~al\mbox{.}(2020)]%
        {carion2020endtoend}
\bibfield{author}{\bibinfo{person}{Nicolas Carion}, \bibinfo{person}{Francisco Massa}, \bibinfo{person}{Gabriel Synnaeve}, \bibinfo{person}{Nicolas Usunier}, \bibinfo{person}{Alexander Kirillov}, {and} \bibinfo{person}{Sergey Zagoruyko}.} \bibinfo{year}{2020}\natexlab{}.
\newblock \bibinfo{title}{End-to-End Object Detection with Transformers}.
\newblock
\newblock
\showeprint[arxiv]{2005.12872}~[cs.CV]


\bibitem[Choi et~al\mbox{.}(2021)]%
        {choi2021deep}
\bibfield{author}{\bibinfo{person}{Kyung~Jun Choi}, \bibinfo{person}{Jung~Eun Choi}, \bibinfo{person}{Hyeon~Cheol Roh}, \bibinfo{person}{Jun~Soo Eun}, \bibinfo{person}{Jong~Min Kim}, \bibinfo{person}{Yong~Kyun Shin}, \bibinfo{person}{Min~Chae Kang}, \bibinfo{person}{Joon~Kyo Chung}, \bibinfo{person}{Chaeyeon Lee}, \bibinfo{person}{Dongyoung Lee}, {et~al\mbox{.}}} \bibinfo{year}{2021}\natexlab{}.
\newblock \showarticletitle{Deep learning models for screening of high myopia using optical coherence tomography}.
\newblock \bibinfo{journal}{\emph{Scientific reports}} \bibinfo{volume}{11}, \bibinfo{number}{1} (\bibinfo{year}{2021}), \bibinfo{pages}{21663}.
\newblock


\bibitem[Devlin et~al\mbox{.}(2018)]%
        {devlin2018bert}
\bibfield{author}{\bibinfo{person}{Jacob Devlin}, \bibinfo{person}{Ming-Wei Chang}, \bibinfo{person}{Kenton Lee}, {and} \bibinfo{person}{Kristina Toutanova}.} \bibinfo{year}{2018}\natexlab{}.
\newblock \showarticletitle{Bert: Pre-training of deep bidirectional transformers for language understanding}.
\newblock \bibinfo{journal}{\emph{arXiv preprint arXiv:1810.04805}} (\bibinfo{year}{2018}).
\newblock


\bibitem[Dosovitskiy et~al\mbox{.}(2020)]%
        {trans001}
\bibfield{author}{\bibinfo{person}{Alexey Dosovitskiy}, \bibinfo{person}{Lucas Beyer}, \bibinfo{person}{Alexander Kolesnikov}, \bibinfo{person}{Dirk Weissenborn}, \bibinfo{person}{Xiaohua Zhai}, \bibinfo{person}{Thomas Unterthiner}, \bibinfo{person}{Mostafa Dehghani}, \bibinfo{person}{Matthias Minderer}, \bibinfo{person}{Georg Heigold}, \bibinfo{person}{Sylvain Gelly}, \bibinfo{person}{Jakob Uszkoreit}, {and} \bibinfo{person}{Neil Houlsby}.} \bibinfo{year}{2020}\natexlab{}.
\newblock \showarticletitle{An Image is Worth 16x16 Words: Transformers for Image Recognition at Scale}.
\newblock \bibinfo{journal}{\emph{CoRR}}  \bibinfo{volume}{abs/2010.11929} (\bibinfo{year}{2020}).
\newblock
\showeprint[arXiv]{2010.11929}
\urldef\tempurl%
\url{https://arxiv.org/abs/2010.11929}
\showURL{%
\tempurl}


\bibitem[Fang et~al\mbox{.}(2022)]%
        {fang2022self}
\bibfield{author}{\bibinfo{person}{Leyuan Fang}, \bibinfo{person}{Jiahuan Guo}, \bibinfo{person}{Xingxin He}, {and} \bibinfo{person}{Muxing Li}.} \bibinfo{year}{2022}\natexlab{}.
\newblock \showarticletitle{Self-supervised patient-specific features learning for OCT image classification}.
\newblock \bibinfo{journal}{\emph{Medical \& Biological Engineering \& Computing}} \bibinfo{volume}{60}, \bibinfo{number}{10} (\bibinfo{year}{2022}), \bibinfo{pages}{2851--2863}.
\newblock


\bibitem[Friberg et~al\mbox{.}(2011)]%
        {friberg2011analysis}
\bibfield{author}{\bibinfo{person}{Thomas~R Friberg}, \bibinfo{person}{Richard~A Bilonick}, {and} \bibinfo{person}{Peter~M Brennen}.} \bibinfo{year}{2011}\natexlab{}.
\newblock \showarticletitle{Analysis of the relationship between drusen size and drusen area in eyes with age-related macular degeneration}.
\newblock \bibinfo{journal}{\emph{Ophthalmic Surgery, Lasers and Imaging Retina}} \bibinfo{volume}{42}, \bibinfo{number}{5} (\bibinfo{year}{2011}), \bibinfo{pages}{369--375}.
\newblock


\bibitem[Gholami et~al\mbox{.}(2023)]%
        {sinagh}
\bibfield{author}{\bibinfo{person}{Sina Gholami}, \bibinfo{person}{Jennifer~I. Lim}, \bibinfo{person}{Theodore Leng}, \bibinfo{person}{Sally Shin~Yee Ong}, \bibinfo{person}{Atalie~Carina Thompson}, {and} \bibinfo{person}{Minhaj~Nur Alam}.} \bibinfo{year}{2023}\natexlab{}.
\newblock \showarticletitle{Federated learning for diagnosis of age-related macular degeneration}.
\newblock \bibinfo{journal}{\emph{Frontiers in Medicine}}  \bibinfo{volume}{10} (\bibinfo{year}{2023}).
\newblock
\showISSN{2296-858X}
\urldef\tempurl%
\url{https://doi.org/10.3389/fmed.2023.1259017}
\showDOI{\tempurl}


\bibitem[He et~al\mbox{.}(2022)]%
        {he2022masked}
\bibfield{author}{\bibinfo{person}{Kaiming He}, \bibinfo{person}{Xinlei Chen}, \bibinfo{person}{Saining Xie}, \bibinfo{person}{Yanghao Li}, \bibinfo{person}{Piotr Doll{\'a}r}, {and} \bibinfo{person}{Ross Girshick}.} \bibinfo{year}{2022}\natexlab{}.
\newblock \showarticletitle{Masked autoencoders are scalable vision learners}. In \bibinfo{booktitle}{\emph{Proceedings of the IEEE/CVF conference on computer vision and pattern recognition}}. \bibinfo{pages}{16000--16009}.
\newblock


\bibitem[He et~al\mbox{.}(2015)]%
        {resnet}
\bibfield{author}{\bibinfo{person}{Kaiming He}, \bibinfo{person}{Xiangyu Zhang}, \bibinfo{person}{Shaoqing Ren}, {and} \bibinfo{person}{Jian Sun}.} \bibinfo{year}{2015}\natexlab{}.
\newblock \showarticletitle{Deep Residual Learning for Image Recognition}.
\newblock \bibinfo{journal}{\emph{CoRR}}  \bibinfo{volume}{abs/1512.03385} (\bibinfo{year}{2015}).
\newblock
\showeprint[arXiv]{1512.03385}
\urldef\tempurl%
\url{http://arxiv.org/abs/1512.03385}
\showURL{%
\tempurl}


\bibitem[Islam(2022)]%
        {islam2022recent}
\bibfield{author}{\bibinfo{person}{Khawar Islam}.} \bibinfo{year}{2022}\natexlab{}.
\newblock \showarticletitle{Recent advances in vision transformer: A survey and outlook of recent work}.
\newblock \bibinfo{journal}{\emph{arXiv preprint arXiv:2203.01536}} (\bibinfo{year}{2022}).
\newblock


\bibitem[Jannat et~al\mbox{.}(2024)]%
        {jannat2024octselfnet}
\bibfield{author}{\bibinfo{person}{Fatema-E Jannat}, \bibinfo{person}{Sina Gholami}, \bibinfo{person}{Minhaj~Nur Alam}, {and} \bibinfo{person}{Hamed Tabkhi}.} \bibinfo{year}{2024}\natexlab{}.
\newblock \bibinfo{title}{OCT-SelfNet: A Self-Supervised Framework with Multi-Modal Datasets for Generalized and Robust Retinal Disease Detection}.
\newblock
\newblock
\showeprint[arxiv]{2401.12344}~[cs.CV]


\bibitem[Jing and Tian(2020)]%
        {jing2020self}
\bibfield{author}{\bibinfo{person}{Longlong Jing} {and} \bibinfo{person}{Yingli Tian}.} \bibinfo{year}{2020}\natexlab{}.
\newblock \showarticletitle{Self-supervised visual feature learning with deep neural networks: A survey}.
\newblock \bibinfo{journal}{\emph{IEEE transactions on pattern analysis and machine intelligence}} \bibinfo{volume}{43}, \bibinfo{number}{11} (\bibinfo{year}{2020}), \bibinfo{pages}{4037--4058}.
\newblock


\bibitem[Kermany et~al\mbox{.}(2018)]%
        {kermany}
\bibfield{author}{\bibinfo{person}{Daniel~S Kermany}, \bibinfo{person}{Michael Goldbaum}, \bibinfo{person}{Wenjia Cai}, \bibinfo{person}{Carolina~CS Valentim}, \bibinfo{person}{Huiying Liang}, \bibinfo{person}{Sally~L Baxter}, \bibinfo{person}{Alex McKeown}, \bibinfo{person}{Ge Yang}, \bibinfo{person}{Xiaokang Wu}, \bibinfo{person}{Fangbing Yan}, {et~al\mbox{.}}} \bibinfo{year}{2018}\natexlab{}.
\newblock \showarticletitle{Identifying medical diagnoses and treatable diseases by image-based deep learning}.
\newblock \bibinfo{journal}{\emph{cell}} \bibinfo{volume}{172}, \bibinfo{number}{5} (\bibinfo{year}{2018}), \bibinfo{pages}{1122--1131}.
\newblock


\bibitem[Kihara et~al\mbox{.}(2022)]%
        {kihara2022detection}
\bibfield{author}{\bibinfo{person}{Yuka Kihara}, \bibinfo{person}{Mengxi Shen}, \bibinfo{person}{Yingying Shi}, \bibinfo{person}{Xiaoshuang Jiang}, \bibinfo{person}{Liang Wang}, \bibinfo{person}{Rita Laiginhas}, \bibinfo{person}{Cancan Lyu}, \bibinfo{person}{Jin Yang}, \bibinfo{person}{Jeremy Liu}, \bibinfo{person}{Rosalyn Morin}, {et~al\mbox{.}}} \bibinfo{year}{2022}\natexlab{}.
\newblock \showarticletitle{Detection of nonexudative macular neovascularization on structural oct images using vision transformers}.
\newblock \bibinfo{journal}{\emph{Ophthalmology Science}} \bibinfo{volume}{2}, \bibinfo{number}{4} (\bibinfo{year}{2022}), \bibinfo{pages}{100197}.
\newblock


\bibitem[Krishnapriya and Karuna(2023)]%
        {krishnapriya2023pre}
\bibfield{author}{\bibinfo{person}{Srigiri Krishnapriya} {and} \bibinfo{person}{Yepuganti Karuna}.} \bibinfo{year}{2023}\natexlab{}.
\newblock \showarticletitle{Pre-trained deep learning models for brain MRI image classification}.
\newblock \bibinfo{journal}{\emph{Frontiers in Human Neuroscience}}  \bibinfo{volume}{17} (\bibinfo{year}{2023}), \bibinfo{pages}{1150120}.
\newblock


\bibitem[Krizhevsky et~al\mbox{.}(2012)]%
        {NIPS2012_c399862d}
\bibfield{author}{\bibinfo{person}{Alex Krizhevsky}, \bibinfo{person}{Ilya Sutskever}, {and} \bibinfo{person}{Geoffrey~E Hinton}.} \bibinfo{year}{2012}\natexlab{}.
\newblock \showarticletitle{ImageNet Classification with Deep Convolutional Neural Networks}. In \bibinfo{booktitle}{\emph{Advances in Neural Information Processing Systems}}, \bibfield{editor}{\bibinfo{person}{F.~Pereira}, \bibinfo{person}{C.J. Burges}, \bibinfo{person}{L.~Bottou}, {and} \bibinfo{person}{K.Q. Weinberger}} (Eds.), Vol.~\bibinfo{volume}{25}. \bibinfo{publisher}{Curran Associates, Inc.}
\newblock
\urldef\tempurl%
\url{https://proceedings.neurips.cc/paper_files/paper/2012/file/c399862d3b9d6b76c8436e924a68c45b-Paper.pdf}
\showURL{%
\tempurl}


\bibitem[Le et~al\mbox{.}(2020)]%
        {10.1167/tvst.9.2.35}
\bibfield{author}{\bibinfo{person}{David Le}, \bibinfo{person}{Minhaj Alam}, \bibinfo{person}{Cham~K. Yao}, \bibinfo{person}{Jennifer~I. Lim}, \bibinfo{person}{Yi-Ting Hsieh}, \bibinfo{person}{Robison V.~P. Chan}, \bibinfo{person}{Devrim Toslak}, {and} \bibinfo{person}{Xincheng Yao}.} \bibinfo{year}{2020}\natexlab{}.
\newblock \showarticletitle{{Transfer Learning for Automated OCTA Detection of Diabetic Retinopathy}}.
\newblock \bibinfo{journal}{\emph{Translational Vision Science \& Technology}} \bibinfo{volume}{9}, \bibinfo{number}{2} (\bibinfo{date}{07} \bibinfo{year}{2020}), \bibinfo{pages}{35--35}.
\newblock
\showISSN{2164-2591}
\urldef\tempurl%
\url{https://doi.org/10.1167/tvst.9.2.35}
\showDOI{\tempurl}
\showeprint{https://arvojournals.org/arvo/content\_public/journal/tvst/938366/i2164-2591-656-1-1965.pdf}


\bibitem[Leandro et~al\mbox{.}(2023)]%
        {leandro2023oct}
\bibfield{author}{\bibinfo{person}{Inferrera Leandro}, \bibinfo{person}{Borsatti Lorenzo}, \bibinfo{person}{Miladinovic Aleksandar}, \bibinfo{person}{Giglio Rosa}, \bibinfo{person}{Accardo Agostino}, {and} \bibinfo{person}{Tognetto Daniele}.} \bibinfo{year}{2023}\natexlab{}.
\newblock \showarticletitle{OCT-based deep-learning models for the identification of retinal key signs}.
\newblock \bibinfo{journal}{\emph{Scientific Reports}} \bibinfo{volume}{13}, \bibinfo{number}{1} (\bibinfo{year}{2023}), \bibinfo{pages}{14628}.
\newblock


\bibitem[Lee et~al\mbox{.}({[n.\,d.]})]%
        {leedeep}
\bibfield{author}{\bibinfo{person}{CS Lee}, \bibinfo{person}{DM Baughman}, {and} \bibinfo{person}{AY Lee}.} \bibinfo{year}{[n.\,d.]}\natexlab{}.
\newblock \bibinfo{title}{Deep learning is effective for the classification of OCT images of normal versus age-related macular degeneration. Ophthalmol Retina. 2017; 1 (4): 322--7}.
\newblock
\newblock


\bibitem[Lee et~al\mbox{.}(2017)]%
        {LEE2017322}
\bibfield{author}{\bibinfo{person}{Cecilia~S. Lee}, \bibinfo{person}{Doug~M. Baughman}, {and} \bibinfo{person}{Aaron~Y. Lee}.} \bibinfo{year}{2017}\natexlab{}.
\newblock \showarticletitle{Deep Learning Is Effective for Classifying Normal versus Age-Related Macular Degeneration OCT Images}.
\newblock \bibinfo{journal}{\emph{Ophthalmology Retina}} \bibinfo{volume}{1}, \bibinfo{number}{4} (\bibinfo{year}{2017}), \bibinfo{pages}{322--327}.
\newblock
\showISSN{2468-6530}
\urldef\tempurl%
\url{https://doi.org/10.1016/j.oret.2016.12.009}
\showDOI{\tempurl}


\bibitem[Li et~al\mbox{.}(2020)]%
        {li2020octa}
\bibfield{author}{\bibinfo{person}{Mingchao Li}, \bibinfo{person}{Kun Huang}, \bibinfo{person}{Qiuzhuo Xu}, \bibinfo{person}{Jiadong Yang}, \bibinfo{person}{Yuhan Zhang}, \bibinfo{person}{Zexuan Ji}, \bibinfo{person}{Keren Xie}, \bibinfo{person}{Songtao Yuan}, \bibinfo{person}{Qinghuai Liu}, {and} \bibinfo{person}{Qiang Chen}.} \bibinfo{year}{2020}\natexlab{}.
\newblock \showarticletitle{OCTA-500: A Retinal Dataset for Optical Coherence Tomography Angiography Study}.
\newblock \bibinfo{journal}{\emph{arXiv e-prints}} (\bibinfo{year}{2020}), \bibinfo{pages}{arXiv--2012}.
\newblock


\bibitem[Liu et~al\mbox{.}(2022)]%
        {liu2022swin}
\bibfield{author}{\bibinfo{person}{Ze Liu}, \bibinfo{person}{Han Hu}, \bibinfo{person}{Yutong Lin}, \bibinfo{person}{Zhuliang Yao}, \bibinfo{person}{Zhenda Xie}, \bibinfo{person}{Yixuan Wei}, \bibinfo{person}{Jia Ning}, \bibinfo{person}{Yue Cao}, \bibinfo{person}{Zheng Zhang}, \bibinfo{person}{Li Dong}, {et~al\mbox{.}}} \bibinfo{year}{2022}\natexlab{}.
\newblock \showarticletitle{Swin transformer v2: Scaling up capacity and resolution}. In \bibinfo{booktitle}{\emph{Proceedings of the IEEE/CVF conference on computer vision and pattern recognition}}. \bibinfo{pages}{12009--12019}.
\newblock


\bibitem[Liu et~al\mbox{.}(2021)]%
        {liu2021swin}
\bibfield{author}{\bibinfo{person}{Ze Liu}, \bibinfo{person}{Yutong Lin}, \bibinfo{person}{Yue Cao}, \bibinfo{person}{Han Hu}, \bibinfo{person}{Yixuan Wei}, \bibinfo{person}{Zheng Zhang}, \bibinfo{person}{Stephen Lin}, {and} \bibinfo{person}{Baining Guo}.} \bibinfo{year}{2021}\natexlab{}.
\newblock \bibinfo{title}{Swin Transformer: Hierarchical Vision Transformer using Shifted Windows}.
\newblock
\newblock
\showeprint[arxiv]{2103.14030}~[cs.CV]


\bibitem[Loshchilov and Hutter(2019)]%
        {adamw}
\bibfield{author}{\bibinfo{person}{Ilya Loshchilov} {and} \bibinfo{person}{Frank Hutter}.} \bibinfo{year}{2019}\natexlab{}.
\newblock \bibinfo{title}{Decoupled Weight Decay Regularization}.
\newblock
\newblock
\showeprint[arxiv]{1711.05101}~[cs.LG]


\bibitem[Lu et~al\mbox{.}(2018)]%
        {lu2018deep}
\bibfield{author}{\bibinfo{person}{Wei Lu}, \bibinfo{person}{Yan Tong}, \bibinfo{person}{Yue Yu}, \bibinfo{person}{Yiqiao Xing}, \bibinfo{person}{Changzheng Chen}, {and} \bibinfo{person}{Yin Shen}.} \bibinfo{year}{2018}\natexlab{}.
\newblock \showarticletitle{Deep learning-based automated classification of multi-categorical abnormalities from optical coherence tomography images}.
\newblock \bibinfo{journal}{\emph{Translational vision science \& technology}} \bibinfo{volume}{7}, \bibinfo{number}{6} (\bibinfo{year}{2018}), \bibinfo{pages}{41--41}.
\newblock


\bibitem[Okolo et~al\mbox{.}(2022)]%
        {okolo2022ievit}
\bibfield{author}{\bibinfo{person}{Gabriel~Iluebe Okolo}, \bibinfo{person}{Stamos Katsigiannis}, {and} \bibinfo{person}{Naeem Ramzan}.} \bibinfo{year}{2022}\natexlab{}.
\newblock \showarticletitle{IEViT: An enhanced vision transformer architecture for chest X-ray image classification}.
\newblock \bibinfo{journal}{\emph{Computer Methods and Programs in Biomedicine}}  \bibinfo{volume}{226} (\bibinfo{year}{2022}), \bibinfo{pages}{107141}.
\newblock


\bibitem[Qiu and Sun(2019)]%
        {qiu2019self}
\bibfield{author}{\bibinfo{person}{Jiaming Qiu} {and} \bibinfo{person}{Yankui Sun}.} \bibinfo{year}{2019}\natexlab{}.
\newblock \showarticletitle{Self-supervised iterative refinement learning for macular OCT volumetric data classification}.
\newblock \bibinfo{journal}{\emph{Computers in biology and medicine}}  \bibinfo{volume}{111} (\bibinfo{year}{2019}), \bibinfo{pages}{103327}.
\newblock


\bibitem[Schmidt-Erfurth et~al\mbox{.}(2018)]%
        {schmidt2018prediction}
\bibfield{author}{\bibinfo{person}{Ursula Schmidt-Erfurth}, \bibinfo{person}{Sebastian~M Waldstein}, \bibinfo{person}{Sophie Klimscha}, \bibinfo{person}{Amir Sadeghipour}, \bibinfo{person}{Xiaofeng Hu}, \bibinfo{person}{Bianca~S Gerendas}, \bibinfo{person}{Aaron Osborne}, {and} \bibinfo{person}{Hrvoje Bogunovi{\'c}}.} \bibinfo{year}{2018}\natexlab{}.
\newblock \showarticletitle{Prediction of individual disease conversion in early AMD using artificial intelligence}.
\newblock \bibinfo{journal}{\emph{Investigative ophthalmology \& visual science}} \bibinfo{volume}{59}, \bibinfo{number}{8} (\bibinfo{year}{2018}), \bibinfo{pages}{3199--3208}.
\newblock


\bibitem[Scott and Bressler(2013)]%
        {scott2013long}
\bibfield{author}{\bibinfo{person}{Adrienne~W Scott} {and} \bibinfo{person}{Susan~B Bressler}.} \bibinfo{year}{2013}\natexlab{}.
\newblock \showarticletitle{Long-term follow-up of vascular endothelial growth factor inhibitor therapy for neovascular age-related macular degeneration}.
\newblock \bibinfo{journal}{\emph{Current opinion in ophthalmology}} \bibinfo{volume}{24}, \bibinfo{number}{3} (\bibinfo{year}{2013}), \bibinfo{pages}{190--196}.
\newblock


\bibitem[Shazia et~al\mbox{.}(2021)]%
        {shazia2021comparative}
\bibfield{author}{\bibinfo{person}{Anis Shazia}, \bibinfo{person}{Tan~Zi Xuan}, \bibinfo{person}{Joon~Huang Chuah}, \bibinfo{person}{Juliana Usman}, \bibinfo{person}{Pengjiang Qian}, {and} \bibinfo{person}{Khin~Wee Lai}.} \bibinfo{year}{2021}\natexlab{}.
\newblock \showarticletitle{A comparative study of multiple neural network for detection of COVID-19 on chest X-ray}.
\newblock \bibinfo{journal}{\emph{EURASIP journal on advances in signal processing}}  \bibinfo{volume}{2021} (\bibinfo{year}{2021}), \bibinfo{pages}{1--16}.
\newblock


\bibitem[Simonyan and Zisserman(2014)]%
        {vgg}
\bibfield{author}{\bibinfo{person}{Karen Simonyan} {and} \bibinfo{person}{Andrew Zisserman}.} \bibinfo{year}{2014}\natexlab{}.
\newblock \showarticletitle{Very deep convolutional networks for large-scale image recognition}.
\newblock \bibinfo{journal}{\emph{arXiv preprint arXiv:1409.1556}} (\bibinfo{year}{2014}).
\newblock


\bibitem[Srinivas et~al\mbox{.}(2022)]%
        {srinivas2022deep}
\bibfield{author}{\bibinfo{person}{Chetana Srinivas}, \bibinfo{person}{Nandini~Prasad KS}, \bibinfo{person}{Mohammed Zakariah}, \bibinfo{person}{Yousef~Ajmi Alothaibi}, \bibinfo{person}{Kamran Shaukat}, \bibinfo{person}{B Partibane}, \bibinfo{person}{Halifa Awal}, {et~al\mbox{.}}} \bibinfo{year}{2022}\natexlab{}.
\newblock \showarticletitle{Deep transfer learning approaches in performance analysis of brain tumor classification using MRI images}.
\newblock \bibinfo{journal}{\emph{Journal of Healthcare Engineering}}  \bibinfo{volume}{2022} (\bibinfo{year}{2022}).
\newblock


\bibitem[Srinivasan et~al\mbox{.}(2014)]%
        {srinivasan}
\bibfield{author}{\bibinfo{person}{Pratul~P Srinivasan}, \bibinfo{person}{Leo~A Kim}, \bibinfo{person}{Priyatham~S Mettu}, \bibinfo{person}{Scott~W Cousins}, \bibinfo{person}{Grant~M Comer}, \bibinfo{person}{Joseph~A Izatt}, {and} \bibinfo{person}{Sina Farsiu}.} \bibinfo{year}{2014}\natexlab{}.
\newblock \showarticletitle{Fully automated detection of diabetic macular edema and dry age-related macular degeneration from optical coherence tomography images}.
\newblock \bibinfo{journal}{\emph{Biomedical optics express}} \bibinfo{volume}{5}, \bibinfo{number}{10} (\bibinfo{year}{2014}), \bibinfo{pages}{3568--3577}.
\newblock


\bibitem[Szegedy et~al\mbox{.}(2015)]%
        {inception}
\bibfield{author}{\bibinfo{person}{Christian Szegedy}, \bibinfo{person}{Wei Liu}, \bibinfo{person}{Yangqing Jia}, \bibinfo{person}{Pierre Sermanet}, \bibinfo{person}{Scott Reed}, \bibinfo{person}{Dragomir Anguelov}, \bibinfo{person}{Dumitru Erhan}, \bibinfo{person}{Vincent Vanhoucke}, {and} \bibinfo{person}{Andrew Rabinovich}.} \bibinfo{year}{2015}\natexlab{}.
\newblock \showarticletitle{Going deeper with convolutions}. In \bibinfo{booktitle}{\emph{Proceedings of the IEEE conference on computer vision and pattern recognition}}. \bibinfo{pages}{1--9}.
\newblock


\bibitem[Tang et~al\mbox{.}(2022)]%
        {tang2022selfsupervised}
\bibfield{author}{\bibinfo{person}{Yucheng Tang}, \bibinfo{person}{Dong Yang}, \bibinfo{person}{Wenqi Li}, \bibinfo{person}{Holger Roth}, \bibinfo{person}{Bennett Landman}, \bibinfo{person}{Daguang Xu}, \bibinfo{person}{Vishwesh Nath}, {and} \bibinfo{person}{Ali Hatamizadeh}.} \bibinfo{year}{2022}\natexlab{}.
\newblock \bibinfo{title}{Self-Supervised Pre-Training of Swin Transformers for 3D Medical Image Analysis}.
\newblock
\newblock
\showeprint[arxiv]{2111.14791}~[cs.CV]


\bibitem[Touvron et~al\mbox{.}(2021)]%
        {touvron2021training}
\bibfield{author}{\bibinfo{person}{Hugo Touvron}, \bibinfo{person}{Matthieu Cord}, \bibinfo{person}{Matthijs Douze}, \bibinfo{person}{Francisco Massa}, \bibinfo{person}{Alexandre Sablayrolles}, {and} \bibinfo{person}{Hervé Jégou}.} \bibinfo{year}{2021}\natexlab{}.
\newblock \bibinfo{title}{Training data-efficient image transformers \& distillation through attention}.
\newblock
\newblock
\showeprint[arxiv]{2012.12877}~[cs.CV]


\bibitem[Tsuji et~al\mbox{.}(2020)]%
        {tsuji2020classification}
\bibfield{author}{\bibinfo{person}{Takumasa Tsuji}, \bibinfo{person}{Yuta Hirose}, \bibinfo{person}{Kohei Fujimori}, \bibinfo{person}{Takuya Hirose}, \bibinfo{person}{Asuka Oyama}, \bibinfo{person}{Yusuke Saikawa}, \bibinfo{person}{Tatsuya Mimura}, \bibinfo{person}{Kenshiro Shiraishi}, \bibinfo{person}{Takenori Kobayashi}, \bibinfo{person}{Atsushi Mizota}, {et~al\mbox{.}}} \bibinfo{year}{2020}\natexlab{}.
\newblock \showarticletitle{Classification of optical coherence tomography images using a capsule network}.
\newblock \bibinfo{journal}{\emph{BMC ophthalmology}} \bibinfo{volume}{20}, \bibinfo{number}{1} (\bibinfo{year}{2020}), \bibinfo{pages}{1--9}.
\newblock


\bibitem[Vaswani et~al\mbox{.}(2017)]%
        {vaswani2017attention}
\bibfield{author}{\bibinfo{person}{Ashish Vaswani}, \bibinfo{person}{Noam Shazeer}, \bibinfo{person}{Niki Parmar}, \bibinfo{person}{Jakob Uszkoreit}, \bibinfo{person}{Llion Jones}, \bibinfo{person}{Aidan~N Gomez}, \bibinfo{person}{{\L}ukasz Kaiser}, {and} \bibinfo{person}{Illia Polosukhin}.} \bibinfo{year}{2017}\natexlab{}.
\newblock \showarticletitle{Attention is all you need}.
\newblock \bibinfo{journal}{\emph{Advances in neural information processing systems}}  \bibinfo{volume}{30} (\bibinfo{year}{2017}).
\newblock


\bibitem[Wang et~al\mbox{.}(2023)]%
        {wang2023visionllm}
\bibfield{author}{\bibinfo{person}{Wenhai Wang}, \bibinfo{person}{Zhe Chen}, \bibinfo{person}{Xiaokang Chen}, \bibinfo{person}{Jiannan Wu}, \bibinfo{person}{Xizhou Zhu}, \bibinfo{person}{Gang Zeng}, \bibinfo{person}{Ping Luo}, \bibinfo{person}{Tong Lu}, \bibinfo{person}{Jie Zhou}, \bibinfo{person}{Yu Qiao}, {and} \bibinfo{person}{Jifeng Dai}.} \bibinfo{year}{2023}\natexlab{}.
\newblock \bibinfo{title}{VisionLLM: Large Language Model is also an Open-Ended Decoder for Vision-Centric Tasks}.
\newblock
\newblock
\showeprint[arxiv]{2305.11175}~[cs.CV]


\bibitem[Wang et~al\mbox{.}(2016)]%
        {wang2016genetic}
\bibfield{author}{\bibinfo{person}{Wenqiu Wang}, \bibinfo{person}{Katarzyna Gawlik}, \bibinfo{person}{Joe Lopez}, \bibinfo{person}{Cindy Wen}, \bibinfo{person}{Jie Zhu}, \bibinfo{person}{Frances Wu}, \bibinfo{person}{William Shi}, \bibinfo{person}{Samuel Scheibler}, \bibinfo{person}{Huimin Cai}, \bibinfo{person}{Ram Vairavan}, {et~al\mbox{.}}} \bibinfo{year}{2016}\natexlab{}.
\newblock \showarticletitle{Genetic and environmental factors strongly influence risk, severity and progression of age-related macular degeneration}.
\newblock \bibinfo{journal}{\emph{Signal transduction and targeted therapy}} \bibinfo{volume}{1}, \bibinfo{number}{1} (\bibinfo{year}{2016}), \bibinfo{pages}{1--6}.
\newblock


\bibitem[Wang et~al\mbox{.}(2022)]%
        {wang2022semi}
\bibfield{author}{\bibinfo{person}{Wei Wang}, \bibinfo{person}{Ran Jiang}, \bibinfo{person}{Ning Cui}, \bibinfo{person}{Qian Li}, \bibinfo{person}{Feng Yuan}, {and} \bibinfo{person}{Zhifeng Xiao}.} \bibinfo{year}{2022}\natexlab{}.
\newblock \showarticletitle{Semi-supervised vision transformer with adaptive token sampling for breast cancer classification}.
\newblock \bibinfo{journal}{\emph{Frontiers in Pharmacology}}  \bibinfo{volume}{13} (\bibinfo{year}{2022}), \bibinfo{pages}{929755}.
\newblock


\bibitem[{World Health Organization}(2023)]%
        {who_blindness_visual_impairment}
\bibfield{author}{\bibinfo{person}{{World Health Organization}}.} \bibinfo{year}{2023}\natexlab{}.
\newblock \bibinfo{booktitle}{\emph{Blindness and visual impairment}}.
\newblock
\urldef\tempurl%
\url{https://www.who.int/news-room/fact-sheets/detail/blindness-and-visual-impairment}
\showURL{%
\tempurl}


\bibitem[Wu et~al\mbox{.}(2021)]%
        {wu2021vision}
\bibfield{author}{\bibinfo{person}{Jianfang Wu}, \bibinfo{person}{Ruo Hu}, \bibinfo{person}{Zhenghong Xiao}, \bibinfo{person}{Jiaxu Chen}, {and} \bibinfo{person}{Jingwei Liu}.} \bibinfo{year}{2021}\natexlab{}.
\newblock \showarticletitle{Vision Transformer-based recognition of diabetic retinopathy grade}.
\newblock \bibinfo{journal}{\emph{Medical Physics}} \bibinfo{volume}{48}, \bibinfo{number}{12} (\bibinfo{year}{2021}), \bibinfo{pages}{7850--7863}.
\newblock


\bibitem[Xie et~al\mbox{.}(2022)]%
        {simmim}
\bibfield{author}{\bibinfo{person}{Zhenda Xie}, \bibinfo{person}{Zheng Zhang}, \bibinfo{person}{Yue Cao}, \bibinfo{person}{Yutong Lin}, \bibinfo{person}{Jianmin Bao}, \bibinfo{person}{Zhuliang Yao}, \bibinfo{person}{Qi Dai}, {and} \bibinfo{person}{Han Hu}.} \bibinfo{year}{2022}\natexlab{}.
\newblock \showarticletitle{Simmim: A simple framework for masked image modeling}. In \bibinfo{booktitle}{\emph{Proceedings of the IEEE/CVF Conference on Computer Vision and Pattern Recognition}}. \bibinfo{pages}{9653--9663}.
\newblock


\bibitem[Yang et~al\mbox{.}(2021)]%
        {yang2021detection}
\bibfield{author}{\bibinfo{person}{Dandi Yang}, \bibinfo{person}{Cristhian Martinez}, \bibinfo{person}{Lara Visu{\~n}a}, \bibinfo{person}{Hardev Khandhar}, \bibinfo{person}{Chintan Bhatt}, {and} \bibinfo{person}{Jesus Carretero}.} \bibinfo{year}{2021}\natexlab{}.
\newblock \showarticletitle{Detection and analysis of COVID-19 in medical images using deep learning techniques}.
\newblock \bibinfo{journal}{\emph{Scientific Reports}} \bibinfo{volume}{11}, \bibinfo{number}{1} (\bibinfo{year}{2021}), \bibinfo{pages}{19638}.
\newblock


\bibitem[Yi et~al\mbox{.}(2009)]%
        {yi2009spectral}
\bibfield{author}{\bibinfo{person}{Kayoung Yi}, \bibinfo{person}{Mircea Mujat}, \bibinfo{person}{Boris~H Park}, \bibinfo{person}{Wei Sun}, \bibinfo{person}{Joan~W Miller}, \bibinfo{person}{Johanna~M Seddon}, \bibinfo{person}{Lucy~H Young}, \bibinfo{person}{Johannes~F de Boer}, {and} \bibinfo{person}{Teresa~C Chen}.} \bibinfo{year}{2009}\natexlab{}.
\newblock \showarticletitle{Spectral domain optical coherence tomography for quantitative evaluation of drusen and associated structural changes in non-neovascular age-related macular degeneration}.
\newblock \bibinfo{journal}{\emph{British Journal of Ophthalmology}} \bibinfo{volume}{93}, \bibinfo{number}{2} (\bibinfo{year}{2009}), \bibinfo{pages}{176--181}.
\newblock


\end{thebibliography}

\end{document}